\documentclass[a4paper,fleqn]{cas-sc}

\usepackage{bbm}
\usepackage{multirow}
\usepackage{booktabs}
\usepackage{amsmath,amssymb,amsfonts}
\usepackage{algorithmic}
\usepackage{graphicx}
\usepackage{subcaption}
\usepackage{textcomp}
\newtheorem{theorem}{Theorem}
\newtheorem{lemma}{Lemma}

\usepackage[authoryear]{natbib}

\def\tsc#1{\csdef{#1}{\textsc{\lowercase{#1}}\xspace}}
\tsc{WGM}
\tsc{QE}

\begin{document}
\let\WriteBookmarks\relax
\def\floatpagepagefraction{1}
\def\textpagefraction{.001}

\shorttitle{FedSemiDG}    

\shortauthors{Z. Deng et. al}



%

\title [mode = title]{FedSemiDG: Domain Generalized Federated Semi-supervised Medical Image Segmentation}



\author[1,2]{Zhipeng Deng}[
orcid=0009-0005-7602-5460
]
\ead[url]{deng.z.aa@m.titech.ac.jp}

\affiliation[1]{%
    organization={Medical Artificial Intelligence Lab, Westlake University},%
    city={Hangzhou},%
    country={China}
}

\affiliation[2]{%
    organization={Department of Information and Communication Engineering, School of Engineering, Institute of Science Tokyo},%
    city={Tokyo},%
    country={Japan}
}

\author[3]{Zhe Xu}[
orcid=0000-0002-1950-0959
]

\ead[url]{jackxz@link.cuhk.edu.hk}

\affiliation[3]{%
    organization={Department of Biomedical Engineering, The Chinese University of Hong Kong},%
    city={Hong Kong},%
    country={China}
}

\author[2]{Tsuyoshi Isshiki}
\ead[url]{isshiki@ict.e.titech.ac.jp}


\author[1]{Yefeng Zheng}[
orcid=0000-0003-2195-2847
]
\ead[url]{zhengyefeng@westlake.edu.cn}
\cormark[1]
\cortext[1]{Corresponding author}



\begin{abstract}
    Medical image segmentation is challenging due to the diversity of medical images and the lack of labeled data, which motivates recent developments in federated semi-supervised learning (FSSL) to leverage a large amount of unlabeled data from multiple centers for model training without sharing raw data. However, what remains under-explored in FSSL is the domain shift problem which may cause suboptimal model aggregation and low effectivity of the utilization of unlabeled data, eventually leading to unsatisfactory performance in unseen domains. In this paper, we explore this previously ignored scenario, namely domain generalized federated semi-supervised learning (FedSemiDG), which aims to learn a model in a distributed manner from multiple domains with limited labeled data and abundant unlabeled data such that the model can generalize well to unseen domains. We present a novel framework, Federated Generalization-Aware Semi-Supervised Learning (FGASL), to address the challenges in FedSemiDG by effectively tackling critical issues at both global and local levels. In our proposed framework, globally, we introduce \textbf{G}eneralization-\textbf{A}ware \textbf{A}ggregation (GAA), assigning adaptive weights to local models based on their generalization performance. Locally, we use a \textbf{D}ual-Teacher Adaptive Pseudo Label \textbf{R}efinement (DR) strategy to combine global and domain-specific knowledge, generating more reliable pseudo labels. Additionally, \textbf{P}erturbation-\textbf{I}nvariant \textbf{A}lignment (PIA) enforces feature consistency under perturbations, promoting domain-invariant learning. Extensive experiments on four medical segmentation tasks (cardiac MRI, spine MRI, bladder cancer MRI and colorectal polyp) demonstrate that our method significantly outperforms state-of-the-art FSSL and domain generalization approaches, achieving robust generalization on unseen domains. This work provides a practical solution for addressing domain shifts in federated semi-supervised learning, advancing multi-center collaboration in privacy-sensitive healthcare applications. The code will be made public upon acceptance.
\end{abstract}




\maketitle




\section{Introduction}
\label{sec:introduction}
Artificial intelligence (AI), particularly deep learning has gradually changed the landscape of computer-aided diagnosis (CAD), including medical image segmentation. The success of deep learning heavily relies on the availability of large-scale datasets. Due to stringent privacy regulations, however, it is difficult to collect large-scale medical image datasets from multiple centers for centralized learning, which hinders the development of medical AI.  As a promising solution to this challenge, federated learning (FL) \citep{mcmahan2017communication} has drawn great attention in the healthcare domain, which enables multi-center collaboration of model training without sharing raw data. 

In the context of medical image segmentation, although extensive research has been conducted on federated learning \citep{jiang2023fair,sheller2019multi,zhang2024federated,luo2023influence}, it is assumed that all the domains have sufficient labeled data, which is not practical in real-world scenarios. In practice, labeled data is often scarce and expensive to obtain for medical image segmentation tasks. To tackle this issue, some recent research has explored the combination of federated learning and semi-supervised segmentation, namely federated semi-supervised segmentation (FSSS) \citep{yang2021federated,wu2023federated,qiu2023federated,ma2024model}. FSSS aims to effectively utilize abundant unlabeled data from multiple domains to improve the model performance. \cite{yang2021federated} were the pioneer in tackling FSSS for COVID-19 lesion segmentation, where they simply applied the centralized semi-supervised learning algorithm FixMatch \citep{sohn2020fixmatch} to the federated setting.  \cite{wu2023federated} tried to improve the performance of FSSS by sharing prototypes among clients, causing potential privacy leakage. \cite{qiu2023federated} proposed to apply Monte Carlo dropout to improve the reliability of pseudo label generation in FSSS, which puts extra computational burden on clients. \cite{ma2024model} considered the scenario where each client holds different structures of local models, where their methods rely on the existence of a shared public dataset in the server, including potential risks of distributional mismatch between public and local datasets and concerns about data accessibility and fairness in real-world applications. 

Another critical challenge in medical image segmentation is domain shift, caused by variations in data collected from different scanners, imaging protocols, or patient populations, which significantly impacts model generalization ability. To overcome the domain shift problem, domain generalization (DG) \citep{li2020domain,li2018deep,chen2024learning,bi2024learning,cheng2025mamba} is introduced to learn models that can generalize well to unseen domains. In centralized learning, some recent research considered both domain shift and insufficient labeled data under the same framework, which was referred to as semi-supervised domain generalization (SemiDG) \citep{yao2022enhancing,wang2023towards,liu2021semi,liu2021disentangled,liu2022vmfnet}. Different from domain generalization (DG) \citep{li2020domain,li2018deep} that puts strong assumptions that all source domains have sufficient labeled data, SemiDG adopts the semi-supervised learning (SSL) paradigm and relaxes this assumption to a more practical scenario where very limited labeled data is available in each source domain while a large amount of data is unlabeled. 

In FL, there are obvious difficulties in training a model that can generalize well to unseen domains in a distributed manner. Federated domain generalization (FedDG) \citep{liu2021feddg,chen2023federated,zhang2023federated,le2024efficiently,raha2025boosting,pourpanah2025federated} is one emerging research area that considers the FL model's generalization ability on the unknown target client with domain shift. However, almost all aforementioned FSSS methods neglect the effect of this domain shift problem \citep{guan2021domain} that may cause suboptimal model aggregation and low effectivity of the utilization of unlabeled data, eventually leading to unsatisfactory generalization performance. Besides, the existing centralized SemiDG methods are not directly applicable to the federated setting due to their reliance on the utilization of multiple domains' data in a centralized manner.

Overall, these studies underscore the need for a new problem setting: domain generalized federated semi-supervised learning (FedSemiDG). This setting addresses a practical scenario in the healthcare domain, where multiple centers collaborate to train a model capable of generalizing to unseen domains in a distributed manner, despite having limited labeled data and abundant unlabeled data.
 
To tackle the domain shift problem in FSSL, several key challenges must be addressed: First, local models trained on diverse domains often exhibit varying levels of generalization, making it essential to design an effective aggregation strategy that ensures the global model achieves robust generalization.  Second, high variances in training samples caused by domain shift may result in low-effectivity of the utilization of unlabeled data, especially for pseudo labeling-based methods \citep{sohn2020fixmatch}. Third, significant variations in image features across domains make learning generalized features in a distributed manner highly challenging, necessitating innovative and effective feature-learning strategies.

In this paper, we introduce the FedSemiDG problem setting and propose a novel framework Federated Generalization-Aware  Semi-Supervised Learning (FGASL) to address this challenge by tackling key issues both globally and locally. On the server side, to mitigate the generalization difference among local models, we assess the generalization ability of each local model by calculating the generalization gap between the local model and the global model on the local dataset. During the model aggregation process, we assign higher weights to local models with larger generalization gaps, aiming to enhance the global model's ability to generalize effectively to underrepresented data. 

On the client side, considering that local models may drift away to client-specific knowledge during local training, we extend the classic teacher-student learning paradigm to a dual-teacher adaptive pseudo label refinement strategy to utilize both global generalized knowledge (static teacher) and local knowledge (dynamic teacher) in the pseudo label generation process. This design is inspired by \cite{xu2024separated}, who proposed leveraging unlabeled data from other centers to enhance the performance of models in a specific center under SSL, demonstrating that shared knowledge across centers can positively impact training. Specifically, the static teacher is a fixed version of the global model, providing better generalized knowledge, while the dynamic teacher, as a moving average of the local model, offers better local knowledge. By combining these two sources, the framework generates more reliable pseudo-labels. At the early stage of training, pseudo labels tend to be noisy due to poor calibration of neural networks \citep{guo2017calibration}. To address this, on top of the dual-teacher learning, we utilize a running uncertainty threshold to filter out the pseudo label with high uncertainty adaptively, which can effectively reduce the noise of pseudo labels and improve the quality of pseudo labels. Furthermore, to enhance robustness against domain-specific variations, we employ a perturbation-invariant alignment strategy. This approach mitigates the risk of overfitting to domain-specific features, ensuring that the model learns more generalized and transferable representations.

Overall, the main contributions of this paper are as follows:

\begin{enumerate}
    \item \textbf{A novel problem setting:} We study a practical yet under-explored scenario in the healthcare domain, namely domain generalized federated semi-supervised learning (FedSemiDG), which aims to learn a model in a distributed manner from multiple domains with limited labeled data and abundant unlabeled data such that the model can generalize well to unseen domains.
    \item \textbf{New insights}: We demonstrate that, under the FedSemiDG setting, either directly applying FSSL methods or simply combining FSSL and DG methods is insufficient to achieve satisfactory performance. 
    \item \textbf{New benchmarks}: We reimplement various state-of-the-art FSSL and DG methods and evaluate them on four commonly used medical image segmentation datasets, which can serve as benchmarks for future research.
    \item \textbf{A new framework:} We propose a novel framework FGASL for FedSemiDG, which tries to solve the domain shift problem in FSSL both globally and locally in three aspects: adaptive model aggregation to get a global model with better generalization ability, dual-teacher adaptive pseudo label refinement to acquire more reliable pseudo labels, and perturbation-invariant alignment to enhance robustness against domain-specific variations.
\end{enumerate}

\section{Related Work}
\subsection{Federated Learning}

Although initially developed for mobile devices, federated learning (FL) has drawn great attention in healthcare due to its potential to enable multi-center collaboration and address the challenges of data privacy and security. FedAvg \citep{mcmahan2017communication} is one of the most popular algorithms for federated learning, which uses a weighted average of the local models to update the global model. The limitations of FedAvg has been addressed in recent works, such as the non-IID data distribution \citep{li2021fedbn,li2020federated,Gao_2022_CVPR,Li_2021_CVPR}, communication efficiency \citep{stich2018local, liconvergence}, privacy and security \citep{bonawitz2017practical,geyer2017differentially,byali2020flash,deng2024enable}.

In the field of medical imaging, FL has drawn increasing interest for allowing the creation of high-performing models from multiple data sources and maintaining privacy at the same time \citep{chen2022personalized,gurler2022federated,deng2024federated,elmas2022federated,guan2024federated}. A notable pilot study demonstrated the feasibility of FL for multi-site brain tumor segmentation \citep{sheller2019multi} without the need to share patient data. Since then, FL has shown great potential in various real-life medical imaging tasks. For instance, FL has been successfully applied to COVID-19 screening \citep{soltan2024scalable}, where clinical data from multiple hospitals were used to improve local model performance. Another example is detecting boundaries of rare cancers  \citep{pati2022federated}, where federated learning enabled significant improvement over a model trained on public datasets.

\subsection{Domain Generalization \& Federated Domain Generalization}
Domain generalization (DG) is a challenging problem in machine learning, which aims to learn a model that can generalize well to unseen target domains without access to any of the target domain data. However, most existing DG methods require access to multiple source domains during training \citep{li2019episodic,li2018domain,xu2021fourier,chen2024learning,cheng2025mamba}, which is often impractical in federated learning (FL) scenarios.	 Single-source domain generalization \citep{li2020domain,huang2020self,choi2023progressive,xu2021robust,wang2021learning} relaxes this requirement by requiring only a single source domain during training. For example, \cite{xu2021robust} applied random convolutions (RC), consisting of a convolution layer randomly initialized for each mini-batch, to enable the model to learn generalizable visual representations by distorting local textures.\cite{choi2023progressive} extended RC to Progressive RC, which recursively stacks random convolution layers with a small kernel size instead of increasing the kernel size. In medical imaging, \cite{li2020domain} proposed learning domain-invariant features by encouraging distribution alignment and low-rank representation.  \cite{bi2024learning} leveraged channel-wise decoupled deep features as queries to guide the learning of generalized representation via cross-attention mechanism. \cite{cheng2025mamba} explored the potential of the Mamba architecture \citep{liu2024vmamba} to address distribution shifts in DG for medical image segmentation.

Nevertheless, there are limited studies addressing the DG problem in FL. While methods like FedBN \citep{li2021fedbn} effectively address domain shifts across a known set of clients by maintaining local batch normalization layers, they are not directly applicable to the FDG setting where the target domain is entirely unseen during training. \cite{liu2021feddg} were the first to introduce the FedDG problem setting and proposed exchanging amplitude information in the frequency domain among clients to augment training data and improve performance.	Similarly, \cite{chen2023federated} proposed extracting and exchanging the overall domain style of local images among all clients. These early works require sharing information among clients, which could lead to potential privacy leakage. Furthermore, \cite{zhang2023federated} introduced a variance reduction regularizer to the original global objective and proposed optimizing this objective by dynamically calibrating model aggregation weights, which is applicable only to fully labeled datasets.  \cite{le2024efficiently} presented a novel normalization method aimed at filtering out domain-specific features and applied a regularizer to encourage the model to capture domain-invariant representations. More recently, \cite{raha2025boosting} demonstrated that employing advanced pre-trained architectures can significantly enhance the generalization ability of federated models to unseen domains, while \cite{pourpanah2025federated} proposed to perform gradient alignment at both the clients and the server to obtain a more generalized aggregated model in an unsupervised manner.

\subsection{Semi-supervised Segmentation \& Semi-supervised Domain Generalization}
Semi-supervised learning (SSL) is a well-established field with two main paradigms: consistency regularization \citep{miyato2018virtual,xie2020unsupervised,xu2023ambiguity} and pseudo-labeling \citep{sohn2020fixmatch,lee2013pseudo,rizve2021defense}.	The basic idea of consistency regularization is to ensure that the model outputs consistent predictions \citep{sohn2020fixmatch} or features \citep{abuduweili2021adaptive} for different views of the same input. Pseudo-labeling, on the other hand, generates either hard labels \citep{sohn2020fixmatch} or soft labels \citep{tarvainen2017mean} for unlabeled data during training. Due to the high cost of labeling, SSL has shown promising progress in medical image segmentation \citep{xu2022all, bai2023bidirectional,xu2022anti}. 

Moreover, there is growing concern about domain shift in SSL. For example, \cite{bai2023bidirectional} applied a bidirectional copy-paste strategy to alleviate the empirical mismatch problem between labeled and unlabeled data. \cite{xu2024separated} proposed separated collaborative learning for semi-supervised prostate segmentation using multi-site unlabeled magnetic resonance imaging (MRI) data to introduce data heterogeneity. \cite{ma2024constructing} referred to SSL with domain shift as mixed-domain semi-supervised medical image segmentation and designed a symmetric guidance training strategy and a random amplitude MixUp module.	

Furthermore, semi-supervised domain generalization (SemiDG) is a new research area that addresses the domain shift problem in SSL. It focuses on a more practical and challenging scenario where the goal is to learn a model that generalizes well to unseen target domains. Early attempts leverage various techniques to address the SemiDG problem, such as meta-learning \citep{liu2021semi} to simulate and handle domain shifts, Fourier transformation \citep{yao2022enhancing} to augment training data with cross-domain features, and compositionality to model robust features across domains \citep{liu2022vmfnet}. However, these methods are not directly applicable to federated settings because they rely on utilizing multiple domains' data in a centralized manner.  In natural images, Zhou et al. \citep{zhou2023semi} applied style transfer to alter the appearance and texture of input images to enhance consistency learning, which may not be applicable to the medical domain due to the characteristics of medical images.	

\begin{figure}[t]
\centerline{\includegraphics[width=0.7\columnwidth]{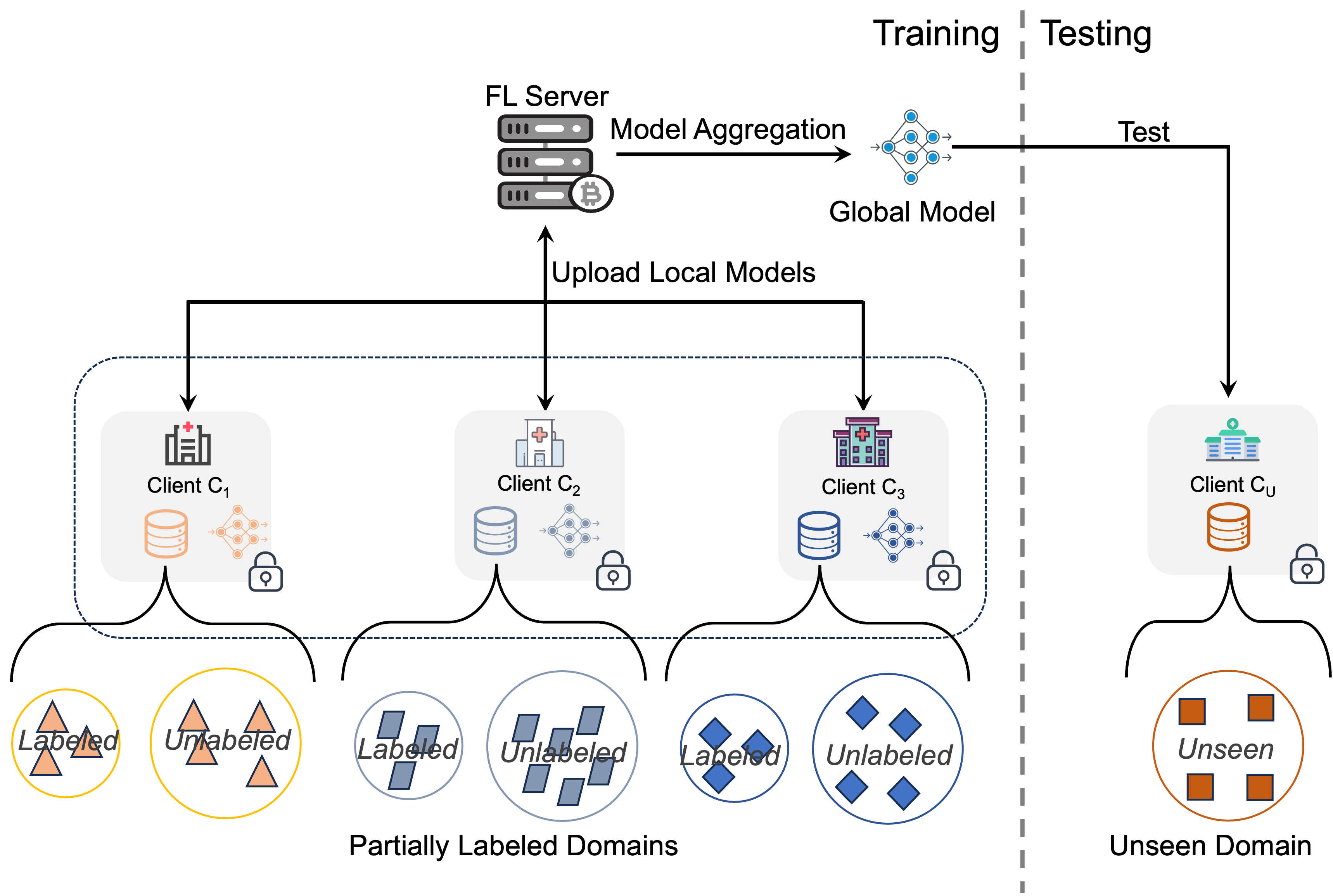}}
\caption{Illustration of domain generalized federated semi-supervised learning (FedSemiDG), where distributed domains collaboratively train a model to generalize to unseen domains.}
\label{fig1}
\end{figure}

\begin{figure*}[!t]
\centerline{\includegraphics[width=0.95\textwidth]{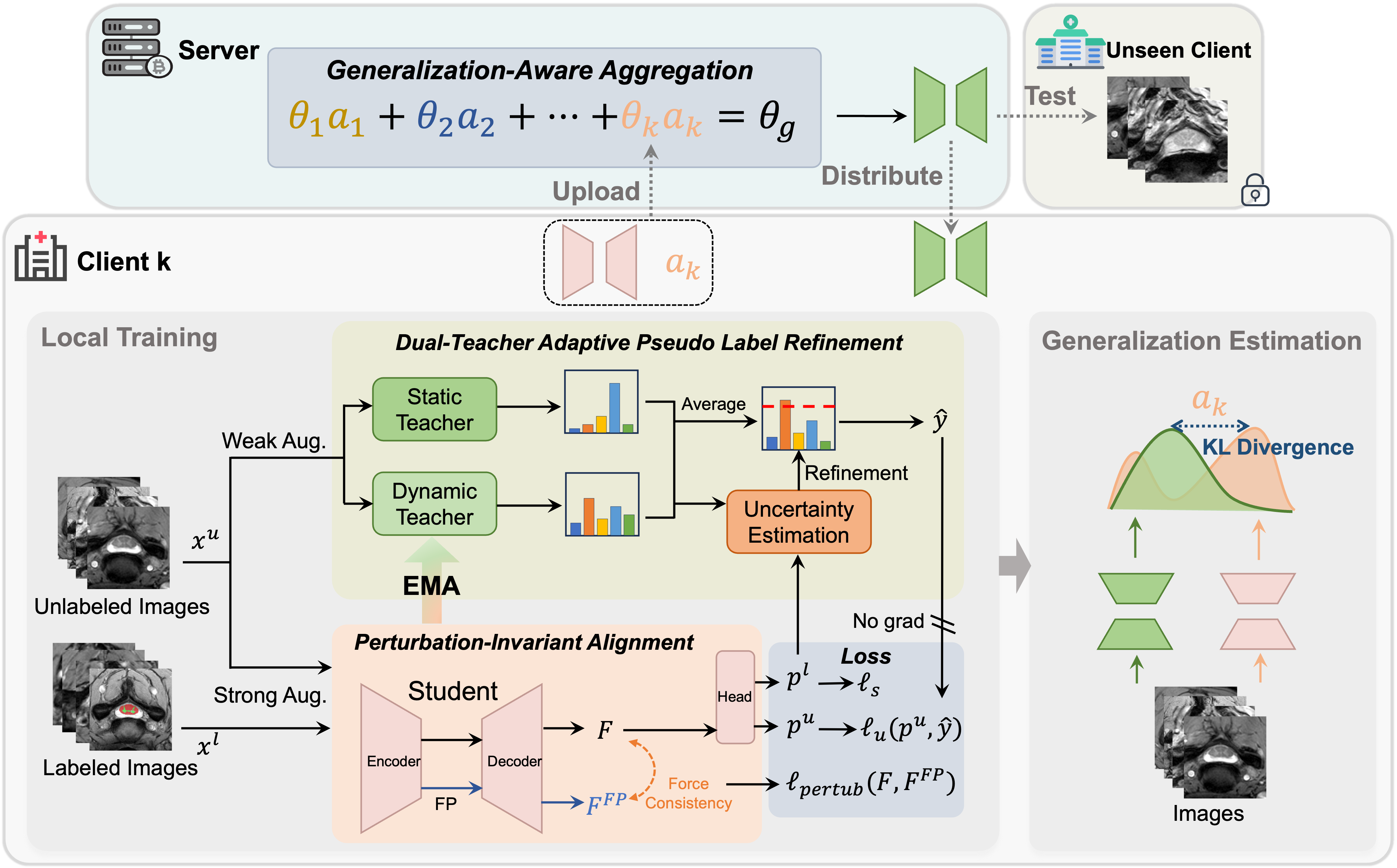}}
\caption{
Overview of the proposed Federated Generalization-Aware Semi-Supervised Learning (FGASL) framework for the FedSemiDG scenario. 
\textbf{(a)} Local training integrates a Dual-Teacher Adaptive Pseudo Label Refinement module, which combines global (static) and local (dynamic) teachers for more reliable pseudo labels with uncertainty-based filtering, and a Perturbation-Invariant Alignment module that enforces consistency between original and perturbed features via feature perturbation (FP). We use channel dropout applied to encoder outputs as FP.
\textbf{(b)} Global aggregation is performed on the server via a generalization-aware weighting strategy, where clients (each corresponding to one domain) contribute their local models weighted by estimated generalization gaps, promoting robust performance on unseen domains.}

\label{fig2}
\end{figure*}

\subsection{Federated Semi-supervised Learning}
Unlike SSL, Federated Semi-supervised Learning (FSSL) \citep{jeong2021federated,liang2022rscfed,liu2024fedcd} integrates the advantages of FL and SSL, leveraging abundant unlabeled data from multiple distributed sources to reduce annotation burdens while simultaneously protecting data privacy, making it particularly attractive in the healthcare domain. For instance,  \cite{yang2021federated} applied FixMatch \citep{sohn2020fixmatch} to generate pseudo labels for unlabeled data in COVID-19 lesion segmentation. \cite{liu2021federated} shared the disease relationship matrix among clients to enforce consistency regularization, a strategy effective for classification tasks but inapplicable to dense prediction due to the lack of global class-level structure. Similarly, \cite{wu2023federated} sought to improve the performance of FSSL by sharing prototypes among labeled and unlabeled clients to encourage the model to learn consistent features. However, sharing additional information among clients may pose risks of privacy leakage. Besides, \cite{jiang2022dynamic} proposed a dynamic bank learning framework (imFed-Semi) to address class imbalance by enforcing sub-bank classification using class prior transitions. This approach relies heavily on image-level pseudo labels and is designed for the scenario where a server holds labeled data while all clients remain fully unlabeled. \cite{qiu2023federated} proposed to apply Monte Carlo dropout to improve the reliability of pseudo label generation in FSSS, which imposes additional computational burdens on clients. \cite{ma2024model} considered a scenario where each client holds differently structured local models. They introduced regularity condensation and regularity fusion to transfer autonomously selected knowledge to ensure personalization, which relies on a shared public dataset on the server. However, this method overlooks that not all tasks have a shared public dataset and that the distribution mismatch between public and local datasets raises potential fairness concerns in real-world applications. 

Compared to existing FSSL works, our study broadens the scope of FSSL and is the first to introduce the Domain Generalized Federated Semi-Supervised Learning (FedSemiDG) problem setting. This extension is motivated by the observation that the domain shift problem has not been adequately considered or addressed in existing FSSL methods, both globally and locally.

\section{Method}
\subsection{Problem Formulation and Framework Overview}
In the FedSemiDG setting,  as depicted in Fig. \ref{fig1}, we consider a central server that coordinates $K$ participants, each holding data from a distinct source domain. Let $\{P^{(k)}(X,Y)\}_{k=1}^K$ denote $K$ such joint distributions over the input space $\mathcal{X}$ and the label space $\mathcal{Y}$. Denote each domain as $D_k$ and the sampled local dataset as $\hat{D}_k$, consisting of a labeled subset $\hat{D}_k^{L} = \{(x_k^l, y_k^l)\}_{i=1}^{N_k^L}$ with $(x_k^l, y_k^l) \sim P^{(k)}(X,Y)$, and an unlabeled subset $\hat{D}_k^{U} = \{x_k^u\}_{i=1}^{N_k^U}$ where $x_k^u \sim P^{(k)}(X)$. We define $\hat{D}_k = \hat{D}_k^{L} \cup \hat{D}_k^{U}$ as the union of the labeled and unlabeled data at the $k$-th participant. $N_k^L$ and $N_k^U$ are the numbers of labeled and unlabeled samples on the $k$-th participant, respectively. The goal of FedSemiDG is to learn a global model $f(x;\theta)$, parameterized by $\theta$, that generalizes well to an unseen target domain $\mathcal{T}$ with a distribution $P^{*}(X,Y)$ which differs from all source distributions $P^{(k)}$ for $k=1, 2, ..., K$. 

Assuming $\mathcal{E}_{\hat{D}_k}(\theta)$ is the local empirical risk minimization objective which incorporates both supervised and unsupervised components $\ell_{s}$ and $\ell_{u}$ balanced by a hyperparameter $\lambda$, and $\mathbf{a} = [a_1, a_2, \ldots, a_K]$ are weights constrained by $\sum_{k=1}^K a_k = 1$ controlling the global optimization process. In FedAvg \citep{mcmahan2017communication}, $a_k$ is proportional to the number of samples in each domain; however, we consider a more general weighting strategy for flexibility. The global optimization problem reduces to the weighted expected risk:
\begin{equation}
    \begin{aligned}
        \min_{\theta} \mathcal{E}_{D}(\theta) &= \sum_{k=1}^{K} a_k \mathcal{E}_{\hat{D}_k}(\theta) \\
        &= \sum_{k=1}^{K} a_k \bigg[ \mathbb{E}_{(x^l,y^l)\sim \widetilde{P}^{(k)}(X,Y)}[\ell_{s}(f(x^l;\theta), y^l)] + \lambda \mathbb{E}_{(x^u)\sim \widetilde{P}^{(k)}(X)}[\ell_{u}(f(x^u;\theta))] \bigg].
    \end{aligned} \label{eq:1}
\end{equation}

As illustrated in Fig.\ref{fig2}, to address the FedSemiDG problem, we propose a novel framework FGASL comprising three key components: generalization-aware aggregation, dual-teacher adaptive pseudo label refinement, and perturbation-invariant alignment. 

\subsection{Generalization-Aware Aggregation}
Due to the domain shift problem, local models trained on different domains may exhibit varied levels of generalization ability. We propose a theoretically-motivated objective function that considers not only the empirical risk but also the consistency of generalization across clients. Formally, as proposed by \cite{zhang2023federated}, we define the \textbf{ideal generalization gap} on a client $k$ as the difference between the global model's performance and the best possible local performance on that client's data:
\begin{equation}
    \mathcal{G}_{\hat D_k}(\theta) := \bigl|\hat{\mathcal E}_{\hat D_k}(\theta) - \hat{\mathcal E}_{\hat D_k}(\theta_k^{*})\bigr|.
    \label{eq:true_gap_def}
\end{equation}
Our high-level optimization goal is to minimize a weighted sum of empirical risks while also minimizing the variance of these ideal gaps across all clients:
\begin{equation}
    \min_{\theta} \mathcal{E}_{D}(\theta) = \sum_{k=1}^{K} a_k \mathcal{E}_{\hat{D}_k}(\theta) + \beta \text{Var}(\{\mathcal{G}_{\hat{D}_k}(\theta)\}_{k=1}^{K}),
    \label{eq:2}
\end{equation}
where $\beta$ is a hyperparameter and $\theta = \sum_{k=1}^{K} a_k \theta_k$ is the global model aggregated with weights $\mathbf{a}$.

However, due to the limited amount of labeled data in our semi-supervised setting, the ideal generalization gap $\mathcal{G}_{\hat D_k}(\theta)$ in Eq.~\eqref{eq:true_gap_def} cannot be computed directly. To overcome this, we introduce a \textbf{practical and computable proxy} for the gap, which is based on the average Kullback--Leibler (KL) divergence between the outputs of the global and local models. We denote this proxy as:
\begin{equation}
    G_{\hat{D}_k}(\theta) := \hat{\mathbb E}_{x\sim\hat D_k}\!\bigl[ \mathrm{KL}\bigl(p_{\theta}(\cdot\mid x)\,\Vert\,p_{\theta_k}(\cdot\mid x)\bigr)\bigr].
    \label{eq:proxy_def}
\end{equation}
The following lemma from our supplementary material provides the crucial theoretical link, showing that our computable proxy $G_{\hat{D}_k}(\theta)$ provides an upper bound for the incomputable ideal gap $\mathcal{G}_{\hat D_k}(\theta)$.

\begin{lemma}[Generalization Gap vs. Predictive KL Proxy]
\label{lem:risk_gap_KL}
Assume the loss function $\ell$ is $L$-Lipschitz (e.g., the Dice loss and the cross-entropy loss) \citep{shalev2014understanding}. The ideal empirical risk gap is bounded by the square root of our KL divergence proxy:
\begin{equation}
    \mathcal{G}_{\hat D_k}(\theta) \;\le\; C_1 \left[ G_{\hat{D}_k}(\theta) \right]^{1/2}.
\end{equation}
\end{lemma}
This theoretical connection justifies using our KL proxy $G_{\hat{D}_k}(\theta)$ to guide the optimization process. Intuitively, a larger proxy value $G_{\hat{D}_k}(\theta)$ suggests a larger underlying ideal gap $\mathcal{G}_{\hat D_k}(\theta)$. By assigning larger aggregation weights to clients with larger proxy values, we aim to improve the global model's generalization ability on under-represented domains. 
Therefore, in our practical algorithm, we optimize a tractable version of Eq.~\eqref{eq:2} by using the variance of our proxy, $\text{Var}(\{G_{\hat{D}_k}(\theta)\}_{k=1}^{K})$, as the regularization term.

Increasing the weight $a_k$ brings the global model closer to the local model $\theta_k$, thereby reducing the generalization gap on the local dataset $\hat{D}_k$. Consequently, the variance of generalization gaps $\text{Var}({G_{\hat{D}_k}(\theta)}_{k=1}^{K})$ can be minimized by assigning higher weights to local models with larger generalization gaps. To achieve this, after local training in the $r$-th FL round, each participant first calculates the generalization gap $G_{\widehat{D}_k}(\theta^r)$ and sends it to the server along with the local model parameters $\theta_k$. This process is inherently privacy-preserving, as the transmitted gap is a single scalar value that contains a high-level summary of model disagreement, revealing minimal information about the local data distribution compared to methods that exchange gradients or data statistics. The server then minimizes the variance of generalization gaps by updating the aggregation weights $\mathbf{a}$ using the following rule:
\begin{equation}
a_k^{r'} = \frac{(G_{\hat{D}_k}(\theta^r) - \mu) \cdot d^r}{\max_j \left(G_{\hat{D}_j}(\theta^r) - \mu \right)} + a_k^{r-1}, \quad
a_k^r = \frac{a_k^{r'}}{\sum_{k=1}^K a_k^{r'}},
\label{eq:5}
\end{equation}
where $\mu = \frac{1}{K} \sum_{k=1}^K G_{\hat{D}_k}(\theta^r)$ and $d^r = (1 - r / R) \cdot d$. Here, $R$ denotes the total number of FL rounds, and $d$ is a hyperparameter controlling the magnitude of the update, similar to $\beta$ in \eqref{eq:2}. The initial weights $a_i^0$ are set to $1/K$ for all participants.

\subsection{Dual-Teacher Adaptive Pseudo Label Refinement}

\cite{xu2023category} utilized the unlabeled data from other centers to support the training of a model in a specific center in SSL and showed promising results, which inspired us to design a dual-teacher adaptive pseudo label refinement strategy to utilize both global generalized knowledge (static teacher $\theta_s$) and local knowledge (dynamic teacher $\theta_d$) in the pseudo label generation process. Specifically, the static teacher is a fixed global model, while the dynamic teacher is initialized from the global model at each FL round and updated by exponential moving average (EMA) of the local model (student model) $\theta_k$: $\theta_d = \pi_1 \theta_d + (1 - \pi_1) \theta_k$, where $\pi_1$ is the EMA decay rate. This design allows the static teacher to mitigate domain-specific drift in the dynamic teacher during local training, thereby producing more reliable pseudo labels.	

To further enhance the robustness of pseudo labels, we adopt a weak-to-strong consistency learning paradigm \citep{sohn2020fixmatch}.  For an unlabeled sample $x$, we first apply weak augmentation $\alpha(\cdot)$ and compute predictions from both teachers.	 The fused probability distribution is defined as: \begin{equation} P_f(c \mid \alpha(x)) = \frac{P_{\theta_s}(c \mid \alpha(x)) + P_{\theta_d}(c \mid \alpha(x))}{2}, \end{equation} from which the pseudo-label is obtained as: \begin{equation} \hat{y} = \arg\max_{c} P_f(c \mid \alpha(x)). \end{equation}

To filter noisy pseudo labels, we compute an adaptive uncertainty threshold $T_{un}$ based on labeled data. At each iteration, the entropy of labeled samples is calculated as $H(x^l) = -\sum_{c=1}^C P_{\theta_k}(c \mid x^l)\log\bigl(P_{\theta_k}(c \mid x^l) + \epsilon\bigr)$, and the threshold $T_{lb}$ is set as the ramp-up quantile $\delta$ of these entropy values. The global threshold $T_{un}$ is then updated via exponential smoothing with decay $\pi_2$: $T_{un}^{(t)} = \pi_2 T_{un}^{(t-1)} + (1 - \pi_2) T_{lb}^{(t)}$. This ensures that the threshold dynamically adapts to the evolving model calibration and complexity during training.	

Next, we apply strong augmentation $A(\cdot)$ to the unlabeled sample $x$ to generate its counterpart $A(x)$ and compute the unlabeled loss $l_u$. For samples that pass the uncertainty threshold ($u_{s+d}(x) \leq T_{un}$), the consistency loss between weakly and strongly augmented views is defined as:
\begin{equation}
    \ell_{u} = \frac{1}{B_u}\sum_{b=1}^{B_u} \mathbbm{1}\bigl( u_{s+d}(x_b^u) \leq T_{un} \bigr) \, L\bigl( \hat{y}_b, P_{\theta_k}(y \mid A(x_b^u)) \bigr),
\end{equation} 
where $B_u$ is the batch size, and $L(\cdot)$ combines cross-entropy loss $l_{ce}$ and Dice loss $l_{dice}$ balanced by a hyperparameter $\eta$ as $L(\cdot) = l_{ce} +  \eta l_{dice}$. The uncertainty $u_{s+d}(x)$ combines entropy from both teachers:
\begin{equation}
u_{s+d}(x_b^u) = -\bigl[ \sum_{c=1}^{C} P_{\theta_s}(c \mid \alpha(x_b^u))\log\bigl(P_{\theta_s}(c \mid \alpha(x_b^u)) + \epsilon\bigr) \\  + \sum_{c=1}^{C} P_{\theta_d}(c \mid \alpha(x_b^u))\log\bigl(P_{\theta_d}(c \mid \alpha(x_b^u)) + \epsilon\bigr) \bigr]/2.
\end{equation} 

This combination of dual-teacher refinement, uncertainty filtering, and consistency learning ensures that only high-quality pseudo labels contribute to model training, effectively leveraging unlabeled data to address domain shifts.

\subsection{Perturbation-Invariant Alignment}
Learning generalized features across multiple domains in a distributed manner is challenging and requires effective strategies. To improve robustness against domain-specific variations, we employ a simple-yet-effective perturbation-invariant alignment strategy. Formally, a segmentation model consists of an encoder $\text{Enc}(\cdot)$, a decoder $\text{Dec}(\cdot)$, and a classification head. We aim to encourage the model to learn features that are invariant to domain-specific perturbations. 

Specifically, we apply a feature perturbation function $\mathcal{P}$ to the encoder output $\text{Enc}(x)$, enforcing the decoder $\text{Dec}(\cdot)$ to generate consistent predictions for both original and perturbed features. In our implementation, $\mathcal{P}$ is instantiated as \textbf{channel-wise dropout}, which randomly zeroes out entire feature channels of $\text{Enc}(x)$ with a fixed probability $p$ (i.e., $\mathcal{P}(\text{Enc}(x)) = \text{DropChannel}(\text{Enc}(x), p)$). This operation is applied at the intersection of the encoder and decoder via the bottleneck and all skip connections, and is designed to encourage the encoder to generate more generic features.

The corresponding loss function is defined as:
\begin{equation}
\ell_{\text{perturb}} = \frac{1}{B_l} \sum_{b=1}^{B_l} \bigl|\bigl| \text{Dec}(\text{Enc}(x_b^l)) - \text{Dec}(\mathcal{P}(\text{Enc}(x_b^l))) \bigr|\bigr|_2^2  +  \frac{1}{B_u} \sum_{b=1}^{B_u} \bigl|\bigl| \text{Dec}(\text{Enc}(x_b^u)) - \text{Dec}(\mathcal{P}(\text{Enc}(x_b^u))) \bigr|\bigr|_2^2,
\end{equation}
where $x_b^l$ and $x_b^u$ represent labeled and unlabeled images within a batch, respectively; $B_l$ and $B_u$ denote the batch sizes for labeled and unlabeled data, respectively; $\mathcal{P}$ is a perturbation function applied to the encoded features; $\text{Enc}(\cdot)$ and $\text{Dec}(\cdot)$ are the encoder and decoder of the segmentation model, respectively. In our experiments, we use channel dropout as the perturbation function. The loss encourages the decoder $\text{Dec}(\cdot)$ to generate consistent outputs for the original and perturbed features, thereby promoting the encoder $\text{Enc}(\cdot)$ to learn domain-invariant representations. More experimental analyses of various perturbation strategies are provided in Sec. \ref{sec:perturb}.

The overall loss is:
$ l = \ell_{s} + \lambda_{1} \ell_{u} + \lambda_{2} \ell_{\text{perturb}},$ where supervised loss $\ell_{s}$ combines the cross-entropy loss and the Dice loss similar to the unsupervised loss $\ell_{u}$, and $\ell_{\text{perturb}}$ is the perturbation-invariant alignment loss, $\lambda_{1}$ and $\lambda_{2}$ are hyperparameters that balance them.

\subsection{Theoretical Justification}
To theoretically ground our framework, we extend the domain generalization theory proposed by \cite{zhang2023federated} to a more practical federated setting where most samples are unlabeled. This analysis culminates in the following generalization bound as in Theorem~\ref{theorem:fgasl_bound}, which motivates the design of our components.  The detailed proof of Theorem~\ref{theorem:fgasl_bound} is provided in the supplementary material. 

\begin{theorem}[Generalization Bound of \textsc{FGASL}]
For any $\delta \in (0,1)$, with probability at least $1-\delta$ over the choice of the training sets $\{\hat{D}_k\}_{k=1}^K$, the generalization error of the global model $\theta$ on an unseen target domain $T$ satisfies:
\begin{equation}
\small
\underbrace{\mathcal{E}_{T}(\theta)-\mathcal{E}_{T}(\theta_T^{*})}_{\text{\footnotesize Generalization\;Error}}
\;\le\;
\sum_{k=1}^{K} a_k
\Bigl(
\underbrace{C_1 \sqrt{G_{\hat{D}_k}(\theta)}}_{\text{\footnotesize Generalization\;Gap}}
+
\underbrace{d_{\mathcal{H}\Delta\mathcal{H}}(\hat{D}_k,T)}_{\text{\footnotesize Domain\,Divergence}}
+
\underbrace{C_2
\frac{\sqrt{\log\frac{K}{\delta}}}{\sqrt{N_k^L}}}_{\text{\footnotesize Estimation\;Error}}
\Bigr)
\;+\;
\underbrace{\lambda}_{\text{\footnotesize Irreducible}}
\end{equation}
where $a_k$ are the GAA weights, $G_{\hat{D}_k}(\theta)$ is the generalization gap's proxy based on the empirical KL divergence, $d_{\mathcal{H}\Delta\mathcal{H}}$ is the domain divergence, and $N_k^L$ is the number of labeled samples. Here, $\theta_T^*$ represents the optimal model for the target domain $T$, while $C_1$ and $C_2$ are theoretical constants related to the Lipschitzness of the loss function and the complexity of the hypothesis class, respectively. Finally, $\lambda$ is the irreducible risk.
\label{theorem:fgasl_bound}
\end{theorem}

Theorem~\ref{theorem:fgasl_bound} reveals that the generalization error is governed by the weighted sum of three key terms. Our FGASL framework is explicitly designed to tighten this bound at two levels:

\noindent \textbf{Global Level (Aggregation)}. Our GAA module directly addresses the aggregation weights $a_k$ to tighten the bound. Based on Lemma~\ref{lem:risk_gap_KL}, we use the computable KL proxy $G_{\hat{D}_k}(\theta)$ to estimate the generalization gap. Specifically, clients with a larger proxy value $G_{\hat{D}_k}(\theta)$ are assigned higher weights. This is a theoretically sound approach because the square root function is monotonic, making the ranking of clients by $G$ equivalent to ranking them by $\sqrt{G}$. This strategy compels the global model to focus on underrepresented domains, aiming to reduce the variance of gaps across clients.

\noindent\textbf{Local Level (Training)}. Our DR and PIA modules are \emph{jointly} designed to tighten the generalization bound. 
The \textbf{Dual-Teacher Refinement (DR)} module directly reduces the generalization gap ($G_{\hat{D}_k}(\theta)$) by regularizing local training against a \emph{frozen} global teacher, preventing the local model from drifting far from the global consensus. 
The \textbf{Perturbation-Invariant Alignment (PIA)} module adds two benefits. First, it also narrows the gap by requiring the local model to produce consistent outputs even when its internal features are perturbed. This discourages over-fitting to narrow, site-specific patterns and encourages more robust representations, keeping the local model’s behaviour aligned with the global model. Second, PIA offers a practical way for a key limitation of FedDG: the lack of direct multi‑domain alignment needed to reduce domain divergence ($d_{\mathcal{H}\Delta\mathcal{H}}$). By promoting invariance to local feature noise, it encourages a natural global alignment toward a common, domain-agnostic feature space, implicitly tightening this second term in the bound.

\section{Experiments}

\begin{table}[ht]
    \centering
    \caption{Dataset statistics across clients. The numbers of image slices are reported in the table.}
    \label{tab:dataset-stats}
    \begin{tabular}{lcccccccc}
    \toprule
    \multicolumn{1}{c}{Dataset} 
    & \multicolumn{2}{c}{Client 1} 
    & \multicolumn{2}{c}{Client 2} 
    & \multicolumn{2}{c}{Client 3} 
    & \multicolumn{2}{c}{Client 4} \\
    \cmidrule(r){2-3} 
    \cmidrule(r){4-5} 
    \cmidrule(r){6-7} 
    \cmidrule(r){8-9}
        & Labeled & Unlabeled 
        & Labeled & Unlabeled 
        & Labeled & Unlabeled 
        & Labeled & Unlabeled \\
    \midrule
    Task 1: Cardiac  & 50  & 1478 & 50  & 1837 & 50  & 984 & 50  & 613 \\
    Task 2: Spine    & 30 & 350 & 30 & 186 & 30 & 339 & 30 & 236 \\
    Task 3: Bladder  & 50  & 815 & 50  & 315 & 50  & 105 & 50  & 139 \\
    Task 4: Polyp   & 30 & 970 & 30 & 582 & 30 & 350 & 30 & 166 \\
    \bottomrule
    \end{tabular} \label{tab:dataset-stats}
    \end{table}

\subsection{Datasets and Evaluation}
\subsubsection{Datasets}
We evaluated our method on four real-world medical image segmentation tasks: cardiac MRI segmentation, spine MRI segmentation, bladder cancer MRI segmentation, and colorectal polyp segmentation.

\textbf{Task 1: Cardiac MRI segmentation.} The dataset \citep{campello2021multi} contains 320 subjects scanned by magnetic resonance scanners from four vendors (i.e., Siemens, Philips, GE, and Canon), primarily for the left ventricle (LV), left ventricle myocardium (MYO), and right ventricle (RV) segmentation tasks. Only the end-systole and end-diastole phases are annotated. The images were center-cropped and resized to 288 $\times$ 288 pixels.

\textbf{Task 2: Spine MRI segmentation.} The dataset \citep{perone2018spinal} was collected from four different medical centers (University College London, Polytechnique Montreal, University of Zurich, and Vanderbilt University) with different MRI systems (Philips Achieva, Siemens Trio, and Siemens Skyra). The spinal cord and gray matter are annotated. The images were center-cropped and resized to 256 $\times$ 256 pixels.

\textbf{Task 3: Bladder cancer MRI segmentation.} The dataset \citep{cao2024multicenter} comprises 275 three-dimensional bladder T2-weighted MRI scans collected from four medical centers. Each scan provides diagnostic pathological labels for muscle invasion and pixel-level annotations of tumor contours. The images were center-cropped and resized to 160 $\times$ 160 pixels.

\textbf{Task 4: Colorectal polyp segmentation.}
We adopted seven widely used colonoscopy image datasets, including Kvasir \citep{kvasir_seg}, CVC-ClinicDB \citep{cvc_clinicdb}, CVC-ColonDB \citep{cvc_colondb}, and ETIS \citep{etis}. These datasets contain images captured from diverse clinical environments, representing different patient populations, endoscopic devices, and lighting conditions. The segmentation targets are polyps of varying sizes and shapes. The images were resized to 352 $\times$ 352 pixels.

Each task consists of \textbf{four} domains: three seen domains and one unseen domain. Each domain was assigned to a unique client, meaning that each client corresponds to one specific data source. Each seen client holds both labeled and unlabeled data from its associated domain. This design simulates real-world federated learning settings in multi-center medical studies, where data heterogeneity arises from different imaging devices, protocols, or patient populations. The detailed statistics of the datasets are shown in Table \ref{tab:dataset-stats}. Specifically, each seen domain contains 50, 30, 50, 30 labeled slices for Task 1, Task 2, Task 3 and Task 4, respectively.  We randomly selected complete 3-D volumes until the required number of labeled slices was reached; all remaining volumes were kept unlabeled.

\subsubsection{Evaluation}
We followed the leave-one-domain-out evaluation protocol for all benchmarks as in \citep{liu2021feddg,zhang2023federated,le2024efficiently}. Specifically, one domain was selected in turn as the unseen target domain, while the remaining domains were used as source domains for training. For evaluation, Tasks 1, 2 and 3 were evaluated with Dice coefficient (DC), Jaccard coefficient (JC), the 95th percentile Hausdorff distance (HD95), and average surface distance (ASD). For Task 4, evaluation was conducted using DC and JC, following common practice \cite{ssformer2022} in previous studies on this dataset. All results were averaged over three independent runs.

\subsection{Implementation Details} \label{sec:imple}
All the experiments were implemented using PyTorch, and all models were trained on an NVIDIA V100 GPU. For each task, to select the optimal hyperparameters, we held out five labeled slices per source domain for validation. A grid search was repeated on three random validation splits, and the most frequently selected best configuration was used for all experiments. For Tasks 1--3, U-Net \citep{ronneberger2015u} was used as the backbone for our method and for all compared methods; for Task 4, SSFormer-L \citep{ssformer2022} was used as the backbone for our method and for all compared methods. The network was optimized by Adam optimizer where the momentum terms were set to 0.9 and 0.99, with the learning rate set to $1 \times 10^{-4}$ for the Task 1, Task 2, and Task 4, and $1 \times 10^{-5}$ for Task 3.  The batch size was 8 for both labeled and unlabeled data. The number of FL rounds was set to 100 and the number of local epochs was set to 1 for all tasks. The hyperparameters to control the trade-off of the loss functions were set to $\lambda_{1} = 1.0$ for all tasks, $\lambda_{2} = 0.3$ for Task 1 and Task 3, and $\lambda_{2} = 0.5$ for Task 2 and Task 4. The EMA decay rate in student-teacher learning was set to $\pi_1 = 0.99$, while for global uncertainty threshold update, $\pi_2 = 0.9$ and the ramp-up quantile $\delta$ was gradually increased from 0.15 to 0.3 in Task 1 and Task 3, and from 0.1 to 0.5 in Task 2 and Task 4. By default, a channel dropout with 30\% probability ($\text{nn.Dropout2d}(\cdot)$ in PyTorch) was adopted for Task 1 and Task 3, and 50\% probability for Task 2 and Task 4 as our feature perturbation, which is inserted at the intersection of the encoder and decoder via the bottleneck and all skip connections. For all tasks, weak data augmentation included random flip, rotation, translation, and scaling. Strong data augmentation included random contrast, brightness, and Gaussian blur.

\subsection{Comparison with State-of-the-Art Methods}

\begin{table}[t]
\centering
\caption{Comparison with state-of-the-art methods on Task 1: Cardiac MRI. The best and second-best results are highlighted in \textbf{bold} and \underline{underlined}, respectively.  Average metrics marked with an asterisk ($^{\ast}$) indicate that the corresponding method is statistically worse than the proposed FGASL ($p<0.05$; see text for details).}
\resizebox{\textwidth}{!}{%
\begin{tabular}{l|c|cccccccc}
\toprule
\multicolumn{10}{c}{Task 1: LV/MYO/RV Segmentation} \\ \midrule
\multirow{2}{*}{Methods} & \multirow{2}{*}{\#L} & \multicolumn{4}{c}{DC (\%) $\uparrow$} & DC (\%)  $\uparrow$ & JC \%  $\uparrow$ & HD95 (mm)$\downarrow$ & ASD (mm) $\downarrow$ \\ \cmidrule{3-10}
& & Vendor A & Vendor B & Vendor C & Vendor D & Avg. & Avg. & Avg. & Avg. \\ \midrule
Local + LabelOnly                         & 50  & 32.81/21.12/22.92 & 48.45/35.07/32.25 & 33.99/18.68/23.84 & 59.94/39.96/31.18 & 33.35$^{\ast}$ & 25.90$^{\ast}$ & 69.56$^{\ast}$ & 43.28$^{\ast}$ \\ 
Local + FixMatch \citep{sohn2020fixmatch} & 50  & 35.62/24.56/23.92 & 53.03/41.83/36.90 & 34.65/21.22/25.48 & 58.05/41.02/31.89 & 35.68$^{\ast}$ & 27.51$^{\ast}$ & 65.08$^{\ast}$ & 38.62$^{\ast}$ \\ 
FL lower bound                            & 150 & 68.74/49.42/40.94 & 77.39/63.94/59.28 & 64.97/50.84/48.54 & 71.86/52.8/57.59  & 58.86$^{\ast}$ & 48.99$^{\ast}$ & 33.49$^{\ast}$ & 28.60$^{\ast}$ \\ \midrule
RSCFed \citep{liang2022rscfed}            & 150 & 68.54/51.8/50.49  & 82.52/73.61/68.65 & 55.12/47.71/52.08 & 68.36/55.72/47.44 & 60.17$^{\ast}$ & 52.29$^{\ast}$ & 32.76$^{\ast}$ & 26.62$^{\ast}$ \\ 
DPL \citep{qiu2023federated}              & 150 & 63.88/56.26/44.65 & 48.86/31.09/17.57 & 44.41/33.48/35.93 & 63.88/56.26/44.65 & 40.16$^{\ast}$ & 30.58$^{\ast}$ & 40.73$^{\ast}$ & 26.09$^{\ast}$ \\ \midrule
FedCD \citep{liu2024fedcd}                & 150 & 62.94/50.06/49.24 & \underline{83.12}/\underline{74.91}/\underline{69.61} & 65.03/56.38/57.33 & 72.77/61.04/50.88 & 62.77$^{\ast}$ & 55.09$^{\ast}$ & 31.43$^{\ast}$ & 25.98$^{\ast}$ \\ 
\,+ RC \citep{xu2021robust}               & 150 & 55.30/46.31/47.60 & 82.66/74.29/70.10 & \textbf{68.95}/\textbf{59.97}/\textbf{59.03} & 73.81/62.06/56.21 & 63.03$^{\ast}$ & 55.52$^{\ast}$ & 31.13$^{\ast}$ & 26.21$^{\ast}$ \\ 
\,+ LDDG \citep{li2020domain}             & 150 & 52.23/35.29/27.51 & 81.41/74.19/71.62 & 67.90/56.27/51.45 & 43.79/29.17/20.36 & 50.93$^{\ast}$ & 43.13$^{\ast}$ & 44.65$^{\ast}$ & 30.78$^{\ast}$ \\ \midrule
FedAvg \citep{mcmahan2017communication} + AugSeq \citep{zhao2023augmentation} & 150 & 58.45/45.91/43.89 & 82.43/73.72/68.8 & 61.84/53.16/56.74 & \underline{79.6}/\underline{67.28}/\textbf{63.30} & 62.93$^{\ast}$ & 55.23$^{\ast}$ & \underline{30.91}$^{\ast}$ & \underline{25.84}$^{\ast}$ \\ 
\,+ RC \citep{xu2021robust}               & 150 & \underline{66.53}/\underline{54.6}/\underline{49.93} & \textbf{83.27}/\textbf{74.94}/\textbf{71.45} & 58.90/49.86/57.27 & 76.15/62.30/57.20 & \underline{63.53}$^{\ast}$ & \underline{56.22}$^{\ast}$ & 31.23$^{\ast}$ & 26.70$^{\ast}$ \\ 
\,+ LDDG \citep{li2020domain}             & 150 & 58.98/47.43/45.86 & 70.02/56.85/47.70 & 61.65/49.74/44.59 & 61.42/41.20/30.90 & 51.36$^{\ast}$ & 42.47$^{\ast}$ & 46.14$^{\ast}$ & 29.73$^{\ast}$ \\ \midrule

FGASL (ours)                             & 150 & \textbf{72.82}/\textbf{60.56}/\textbf{55.92} & 82.08/74.18/70.76 & \underline{66.47}/\underline{57.37}/\underline{59.84} & \textbf{80.51}/\textbf{68.46}/\underline{62.25} & \textbf{67.60} & \textbf{59.76} & \textbf{26.65} & \textbf{21.86} \\ \midrule
FL upper bound                           & -   & 85.07/72.56/64.31 & 89.42/82.29/74.58 & 88.58/80.36/76.88 & 89.42/81.65/80.88 & 80.50 & 72.25 & 13.09 & 9.92 \\ 
\bottomrule
\end{tabular}}
\label{tab:1}
\end{table}

\begin{table}[ht]
\centering
\caption{ Comparison with the state-of-the-art methods on Task 2: Spine MRI Segmentation. The best and second-best results are highlighted in \textbf{bold} and \underline{underlined}, respectively.  Average metrics marked with an asterisk ($^{\ast}$) indicate that the corresponding method is statistically worse than the proposed FGASL ($p<0.05$; see text for details).}
\resizebox{\columnwidth}{!}{%
\begin{tabular}{l|c|cccccccc}
\toprule
\multicolumn{10}{c}{Task 2: Spinal Cord/Grey Matter Segmentation} \\ \midrule
\multirow{2}{*}{Methods} & \multirow{2}{*}{\#L} & \multicolumn{4}{c}{DC (\%) $\uparrow$} & DC (\%) $\uparrow$ & JC (mm) $\uparrow$ & HD95 (mm)$\downarrow$ & ASD $\downarrow$ \\ \cmidrule{3-10}
& & Center A & Center B & Center C & Center D & Avg. & Avg. & Avg. & Avg. \\ \midrule
Local + LabelOnly & 30 & 78.11/51.65 & 63.96/41.16 & 63.38/39.60 & 77.69/58.86 & 59.30$^{\ast}$ & 50.77$^{\ast}$ & 42.15$^{\ast}$ & 25.05$^{\ast}$ \\ 
Local + FixMatch \citep{sohn2020fixmatch} & 30 & 69.41/42.39 & 72.59/43.65 & 60.91/28.08 & 83.69/61.36 & 57.76$^{\ast}$ & 48.15$^{\ast}$ & 45.65$^{\ast}$ & 24.10$^{\ast}$ \\ 
FL lower bound & 90 & 96.01/79.07 & \underline{90.00}/\underline{71.71} & 66.33/40.27 & 95.72/81.19 & 77.54$^{\ast}$ & 68.36$^{\ast}$ & 11.89$^{\ast}$ & 5.81$^{\ast}$ \\ \midrule
RSCFed \citep{liang2022rscfed} & 90 & \textbf{97.01}/\textbf{81.55} & 81.98/55.84 & 71.04/60.05 & \textbf{96.98}/85.61 & 78.76$^{\ast}$ & 70.04$^{\ast}$ & 9.27$^{\ast}$ & 4.00$^{\ast}$ \\
DPL \citep{qiu2023federated} & 90  & 93.06/63.15 & 14.47/3.21 & 20.98/4.03 & 90.28/65.71 & 44.36$^{\ast}$ & 36.20$^{\ast}$ & 79.91$^{\ast}$ & 42.69$^{\ast}$ \\ \midrule
FedCD \citep{liu2024fedcd} & 90  & 95.42/79.53 & 80.40/58.20 & 80.81/64.48 & 95.98/86.22 & 80.13$^{\ast}$ & 70.82$^{\ast}$ & 9.30$^{\ast}$ & 3.89$^{\ast}$ \\ 
+ RC \citep{xu2021robust} & 90 & \underline{96.12}/\underline{80.37} & 88.22/69.88 & 73.59/59.63 & 96.00/85.82 & \underline{81.20}$^{\ast}$ & \underline{72.49}$^{\ast}$ & \underline{7.88}$^{\ast}$ & \underline{3.46}$^{\ast}$ \\ 
+ LDDG \citep{li2020domain} & 90 & 93.60/77.02 & 86.65/53.56 & 78.48/56.55 & 94.90/80.83 & 77.65$^{\ast}$ & 67.62$^{\ast}$ & 10.75$^{\ast}$ & 3.79$^{\ast}$ \\ \midrule
FedAvg \citep{mcmahan2017communication} + AugSeq \citep{zhao2023augmentation} & 90 & 86.83/64.70 & 80.64/57.87 & \underline{81.71}/\textbf{67.38} & 93.34/82.84 & 79.07$^{\ast}$ & 69.28$^{\ast}$ & 11.17$^{\ast}$ & 4.41$^{\ast}$ \\ 
+ RC \citep{xu2021robust} & 90 & 95.13/76.25 & 81.63/63.09 & 73.88/56.17 & 92.43/82.30 & 77.62$^{\ast}$ & 68.03$^{\ast}$ & 11.02$^{\ast}$ & 5.30$^{\ast}$ \\ 
+ LDDG \citep{li2020domain} & 90 & 93.60/77.02 & 79.77/53.57 & 72.26/58.68 & 94.85/75.02 & 77.69$^{\ast}$ & 67.53$^{\ast}$ & 12.13$^{\ast}$ & 5.30$^{\ast}$ \\ \midrule
FGASL (ours) & 90 & 95.48/79.86 & \textbf{95.59}/\textbf{80.68} & \textbf{82.08}/\underline{66.70} & \underline{96.57}/\textbf{86.29} & \textbf{85.41} & \textbf{77.04} & \textbf{4.99} & \textbf{1.96} \\ \midrule
FL upper bound & - & 96.48/81.82 & 94.27/80.51 & 85.29/68.54 & 98.38/86.92 & 86.23 & 77.85 & 7.15 & 2.44 \\ \bottomrule
\end{tabular}%
}
\label{tab:2}
\end{table}

\begin{table}[t]
\centering
\caption{ Comparison with the state-of-the-art methods on Task 3: Bladder Cancer Segmentation. The best and second-best results are highlighted in \textbf{bold} and \underline{underlined}, respectively.  Average metrics marked with an asterisk ($^{\ast}$) indicate that the corresponding method is statistically worse than the proposed FGASL ($p<0.05$; see text for details).}
\resizebox{\columnwidth}{!}{%
\begin{tabular}{l|c|cccccccc}
\toprule
\multicolumn{10}{c}{Task 3: Bladder Cancer Segmentation} \\ \midrule
\multirow{2}{*}{Methods} & \multirow{2}{*}{\#L} & \multicolumn{4}{c}{DC (\%)$\uparrow$} & DC (\%) $\uparrow$ & JC (\%) $\uparrow$ & HD95 (mm)$\downarrow$ & ASD (mm) $\downarrow$ \\ 
\cmidrule{3-10}
& & Center 1 & Center 2 & Center 3 & Center 4 & Avg. & Avg. & Avg. & Avg. \\ \midrule
Local + LabelOnly & 50 & 48.13 & 50.97 & 55.46 & 54.97 & 52.38$^{\ast}$ & 39.76$^{\ast}$ & 44.85$^{\ast}$ & 17.00$^{\ast}$ \\ 
Local + FixMatch \citep{sohn2020fixmatch} & 50 & 51.75 & 61.31 & 53.20 & 48.52 & 53.70$^{\ast}$ & 40.86$^{\ast}$ & 41.85$^{\ast}$ & 17.24$^{\ast}$ \\ 
FL lower bound & 150 & 58.36 & 56.28 & 60.55 & 57.75 & 58.23$^{\ast}$ & 44.55$^{\ast}$ & 58.24$^{\ast}$ & 21.85$^{\ast}$ \\ \midrule
RSCFed \citep{liang2022rscfed} & 150 & 56.77 & 63.68 & 57.20 & 59.61 & 59.31$^{\ast}$ & 46.27$^{\ast}$ & 35.79$^{\ast}$ & 14.21$^{\ast}$ \\ 
DPL \citep{qiu2023federated} & 150 & 55.39 & 52.79 & 57.77 & 59.32 & 56.32$^{\ast}$ & 43.54$^{\ast}$ & 43.69$^{\ast}$ & 15.11$^{\ast}$ \\ \midrule
FedCD \citep{liu2024fedcd} & 150 & 57.97 & 58.02 & 58.12 & 59.84 & 58.49$^{\ast}$ & 45.60$^{\ast}$ & \underline{33.44} & 13.19 \\ 
+ RC \citep{xu2021robust} & 150 & \textbf{60.30} & 54.69 & \underline{61.65} & 59.89 & 59.13$^{\ast}$ & \underline{46.57}$^{\ast}$ & 41.44$^{\ast}$ & 13.83$^{\ast}$ \\ 
+ LDDG \citep{li2020domain} & 150 & 54.80 & 63.86 & 59.08 & 58.46 & 59.05$^{\ast}$ & 46.15$^{\ast}$ & 33.79 & \underline{13.09} \\ \midrule
FedAvg \citep{mcmahan2017communication} + AugSeq \citep{zhao2023augmentation} & 90 & \underline{59.10} & 54.12 & 61.14 & \textbf{60.83} & 58.80$^{\ast}$ & 45.84$^{\ast}$ & 44.42$^{\ast}$ & 15.69$^{\ast}$ \\ 
+ RC \citep{xu2021robust} & 150 & 57.62 & \underline{65.29} & 59.14 & 55.81 & \underline{59.47}$^{\ast}$ & 46.48$^{\ast}$ & 34.80$^{\ast}$ & 13.55$^{\ast}$ \\ 
+ LDDG \citep{li2020domain} & 150 & 21.06 & 49.47 & 53.35 & 56.94 & 45.03$^{\ast}$ & 39.69$^{\ast}$ & 43.89$^{\ast}$ & 19.87$^{\ast}$ \\ \midrule
FGASL (ours) & 150 & 59.10 & \textbf{67.63} & \textbf{62.53} & \underline{60.34} & \textbf{62.40} & \textbf{49.18} & \textbf{33.21} & \textbf{12.13} \\ \midrule
FL upper bound & - & 64.39 & 70.08 & 68.60 & 70.56 & 68.41 & 56.04 & 26.37 & 9.90 \\ 
\bottomrule
\end{tabular}%
}
\label{tab:3}
\end{table}

\begin{table}[h]
\centering
\caption{Comparison with the state-of-the-art methods on Task 4: Colorectal Polyp Segmentation. The best and second-best results are highlighted in \textbf{bold} and \underline{underlined}, respectively. Average metrics marked with an asterisk ($^{\ast}$) indicate that the corresponding method is statistically worse than the proposed FGASL ($p<0.05$).}
\resizebox{\columnwidth}{!}{%
\begin{tabular}{l|c|cccccc}
\toprule
\multicolumn{8}{c}{Task 4: Colorectal Polyp Segmentation} \\ \midrule
\multirow{2}{*}{Methods} & \multirow{2}{*}{\#L} & \multicolumn{4}{c}{DC (\%)$\uparrow$} & DC (\%)$\uparrow$ & JC (\%)$\uparrow$ \\
\cmidrule{3-8}
& & Kvasir & ClinicDB & ColonDB & ETIS & Avg. & Avg. \\ \midrule
Local + LabelOnly & 30 & 75.21 & 75.27 & 61.30 & 55.26 & 66.76$^{\ast}$ & 57.49$^{\ast}$ \\
Local + FixMatch \citep{sohn2020fixmatch} & 30 & 79.34 & 77.45 & 64.54 & 62.18 & 70.88$^{\ast}$ & 61.46$^{\ast}$ \\
FL lower bound & 90 & 58.36 & 56.28 & 60.55 & 57.75 & 58.23$^{\ast}$ & 44.55$^{\ast}$ \\ \midrule
RSCFed \citep{liang2022rscfed} & 90 & 83.87 & 80.75 & 70.73 & 62.77 & 74.78$^{\ast}$ & 66.40$^{\ast}$ \\
DPL \citep{qiu2023federated} & 90 & 82.72 & 80.00 & 66.25 & 58.00 & 71.74$^{\ast}$ & 63.31$^{\ast}$ \\ \midrule
FedCD \citep{liu2024fedcd} & 90 & 84.57 & 82.01 & 70.94 & 64.68 & 75.55$^{\ast}$ & 67.16$^{\ast}$ \\
+ RC \citep{xu2021robust} & 90 & 84.36 & 83.20 & 70.31 & 63.33 & 75.30$^{\ast}$ & 67.03$^{\ast}$ \\
+ LDDG \citep{li2020domain} & 90 & \underline{85.70} & \underline{83.70} & 69.70 & 66.42 & 76.10$^{\ast}$ & 68.01$^{\ast}$ \\ \midrule
FedAvg \citep{mcmahan2017communication} + AugSeq \citep{zhao2023augmentation} & 90 & 85.24 & 82.75 & 72.27 & 65.02 & 76.32$^{\ast}$ & 67.86$^{\ast}$ \\
+ RC \citep{xu2021robust} & 90 & 84.71 & 83.18 & \underline{73.44} & 64.81 & 76.53$^{\ast}$ & 68.07$^{\ast}$ \\
+ LDDG \citep{li2020domain} & 90 & \underline{85.70} & \underline{83.70} & 72.63 & \underline{68.95} & \underline{77.74}$^{\ast}$ & \underline{69.49} \\ \midrule
FGASL (ours) & 90 & \textbf{86.41} & \textbf{85.73} & \textbf{75.27} & \textbf{70.21} & \textbf{79.41} & \textbf{71.25} \\ \midrule
FL upper bound & - & 87.63 & 88.57 & 80.62 & 79.61 & 84.11 & 76.62 \\
\bottomrule
\end{tabular}%
}
\label{tab:4}
\end{table}

\begin{figure}[t!]
\centerline{\includegraphics[width=0.8\columnwidth]{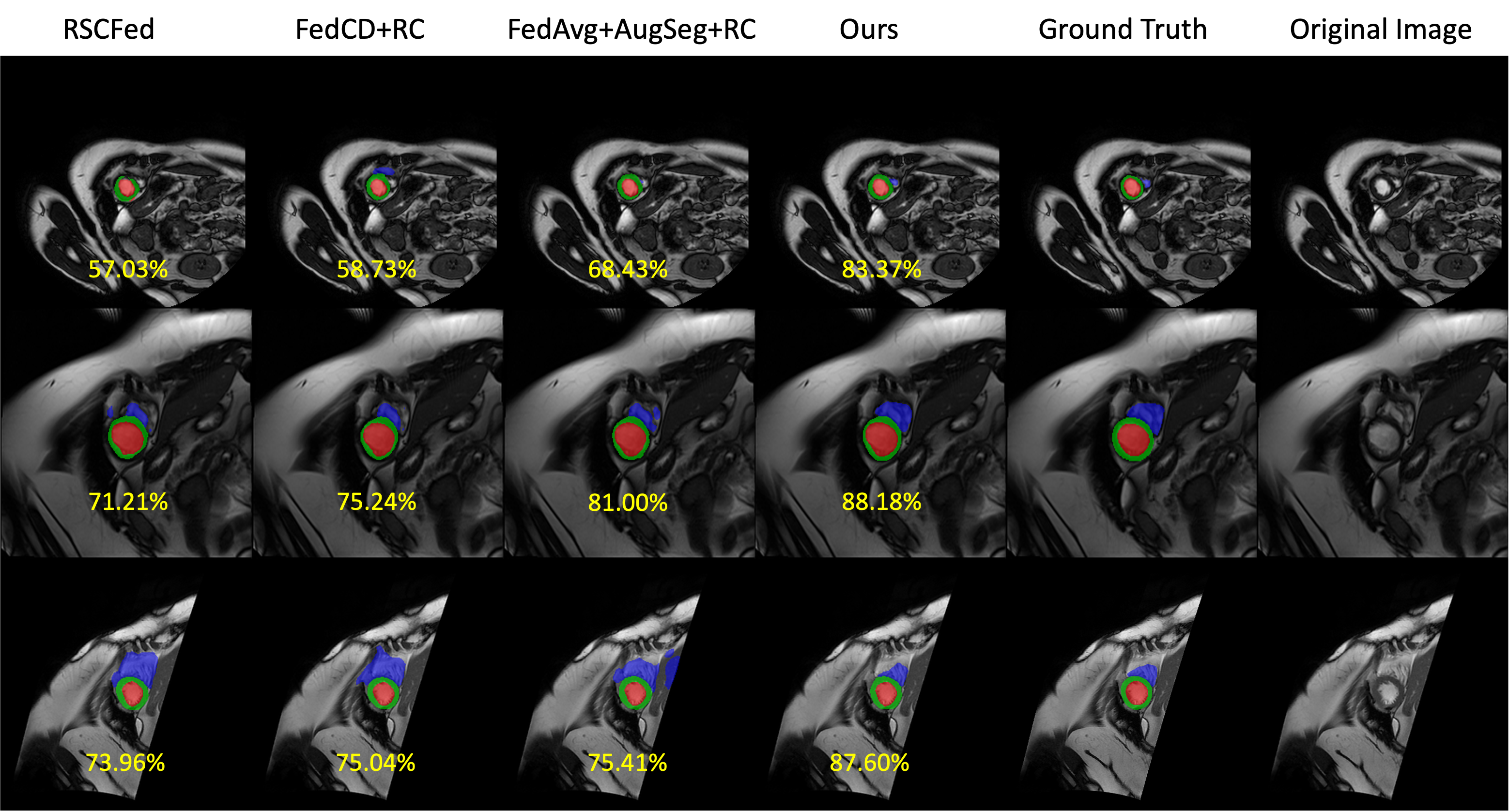}}
\caption{Exemplar cardiac MRI segmentation results on the unseen domain. The Dice scores (\%) are displayed at the bottom.}\label{fig3}
\end{figure}
\begin{figure}[t!]
\centerline{\includegraphics[width=0.8\columnwidth]{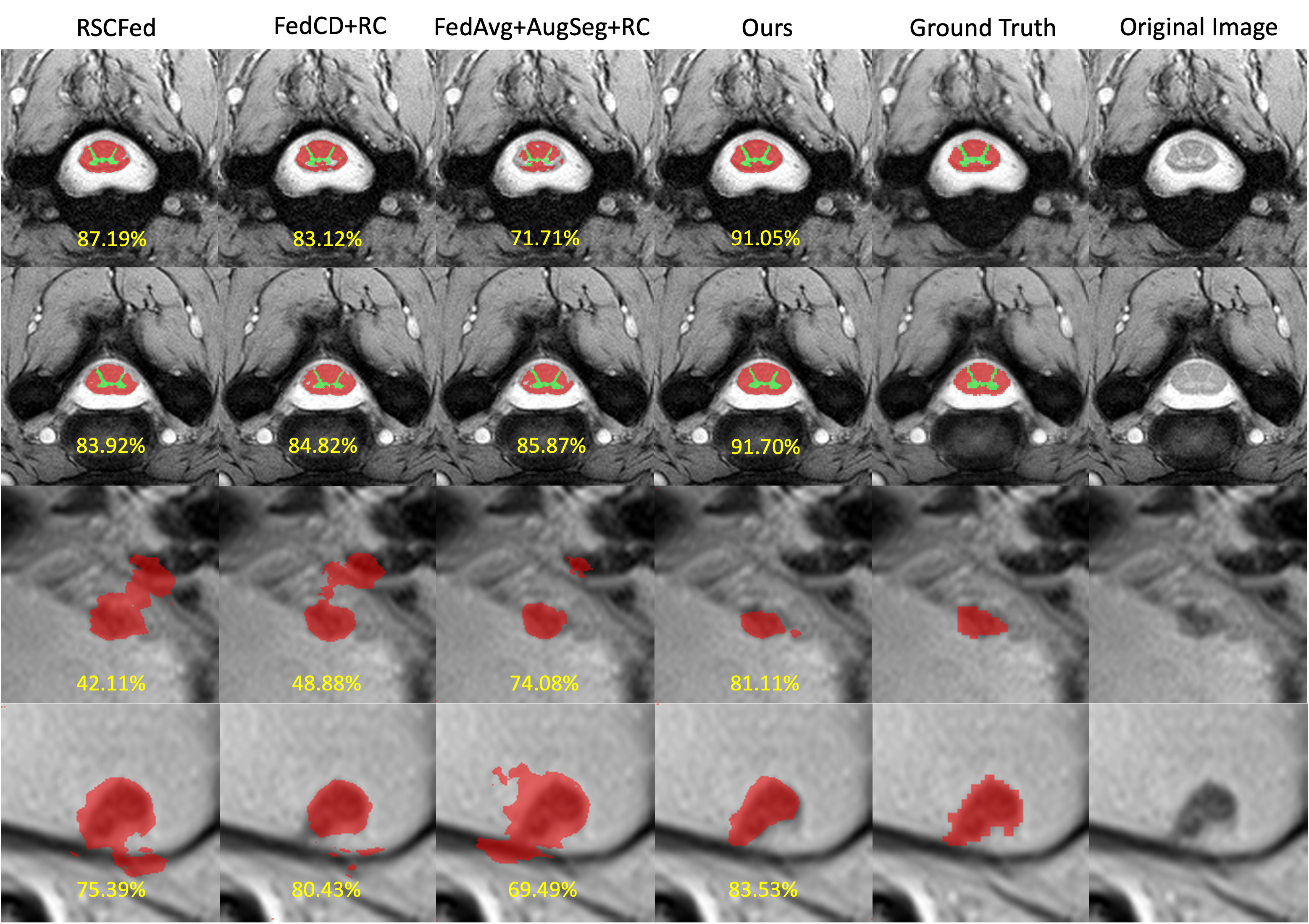}}
\caption{Exemplar segmentation on the unseen domain for spine (first and second rows), and bladder cancer (third and fourth rows). The Dice scores (\%) are displayed at the bottom.}\label{fig4}
\end{figure}

We compared our method with recent state-of-the-art (SOTA) FSSL methods, including RSCFed \citep{liang2022rscfed}, DPL \citep{qiu2023federated}, and FedCD \citep{liu2024fedcd}. To further demonstrate the effectiveness of our approach, we also evaluated FedCD combined with single domain generalization (DG) techniques applicable to the FSSL setting, such as RC \citep{xu2021robust} and LDDG \citep{li2020domain}. Additionally, we combined FedAvg \citep{mcmahan2017communication} with a SOTA semi-supervised learning (SSL) method, AugSeq \citep{zhao2023augmentation}, and further incorporated RC and LDDG for comparison. Moreover, we included several baselines in our evaluation: the FL lower bound, represented by FedAvg with only labeled data; the FL upper bound, represented by FedAvg with fully labeled data; and local training baselines, either using only labeled data or leveraging FixMatch \citep{sohn2020fixmatch}.

Across all four tasks, our method consistently showed significant improvements over the FL lower bound and outperformed existing SOTA approaches or their combinations with DG techniques in terms of average DC, JC, HD95, and ASD, as shown in Tables \ref{tab:1}, \ref{tab:2}, \ref{tab:3}, and \ref{tab:4}. From the four tables, we can observe that FSSL only brings marginal improvements over the FL lower bound, indicating the challenges of learning from cross-domain data in the FSSL setting. Moreover, the DG techniques, such as RC and LDDG, can only slightly improve the performance or even degrade it in some cases. In contrast, our method achieved significant performance gains over the FL lower bound as well as existing SOTA methods. When compared with the second-best method, our approach achieves an improvement of 4.07\%, 4.21\%, 2.93\% and 1.67\% in average DC on the Cardiac MRI, Spine MRI, Bladder Cancer, and Colorectal Polyp Segmentation tasks, respectively. We conducted paired \textit{t}-tests between FGASL and \emph{each} competing method on the three tasks for all average metrics. In most cases the differences were statistically significant ($p<0.05$); for instance, against the second-best approach for average DC the $p$-values were $6.4\times10^{-3}$ (Task 1), $5.3\times10^{-4}$ (Task 2), $3.0\times10^{-3}$ (Task 3) and $6.3\times10^{-3}$ (Task 4). Besides, as shown in Figs. \ref{fig3} and \ref{fig4}, we provide qualitative comparisons on unseen domains in the Cardiac MRI, Spine MRI, and Bladder Cancer Segmentation tasks, where our method consistently achieved more accurate segmentation results compared to some top-performing approaches, demonstrating the robustness of our method in handling domain shift.

To further show that the FedSemiDG setting is clinically meaningful, we compared our method with the local training baselines. Localized training with SSL method FixMatch \citep{sohn2020fixmatch} could not achieve satisfactory performance due to the lack of access to multi-center data. However, our approach showed substantial improvements, demonstrating the effectiveness of leveraging federated frameworks in multi-center collaborative learning. For instance, our method improved Local+FixMatch by 31.92\%, 27.65\%, 8.70\% and 8.53\% in average DC on the Cardiac MRI, Spine MRI, Bladder Cancer and Colorectal Polyp Segmentation tasks, respectively. These results underscore the importance of collaborative learning across distributed clients, especially when labeled data is scarce.

\begin{figure}[t]
    \centering
    \begin{subfigure}[b]{0.32\textwidth}
        \centering
        \includegraphics[width=\textwidth]{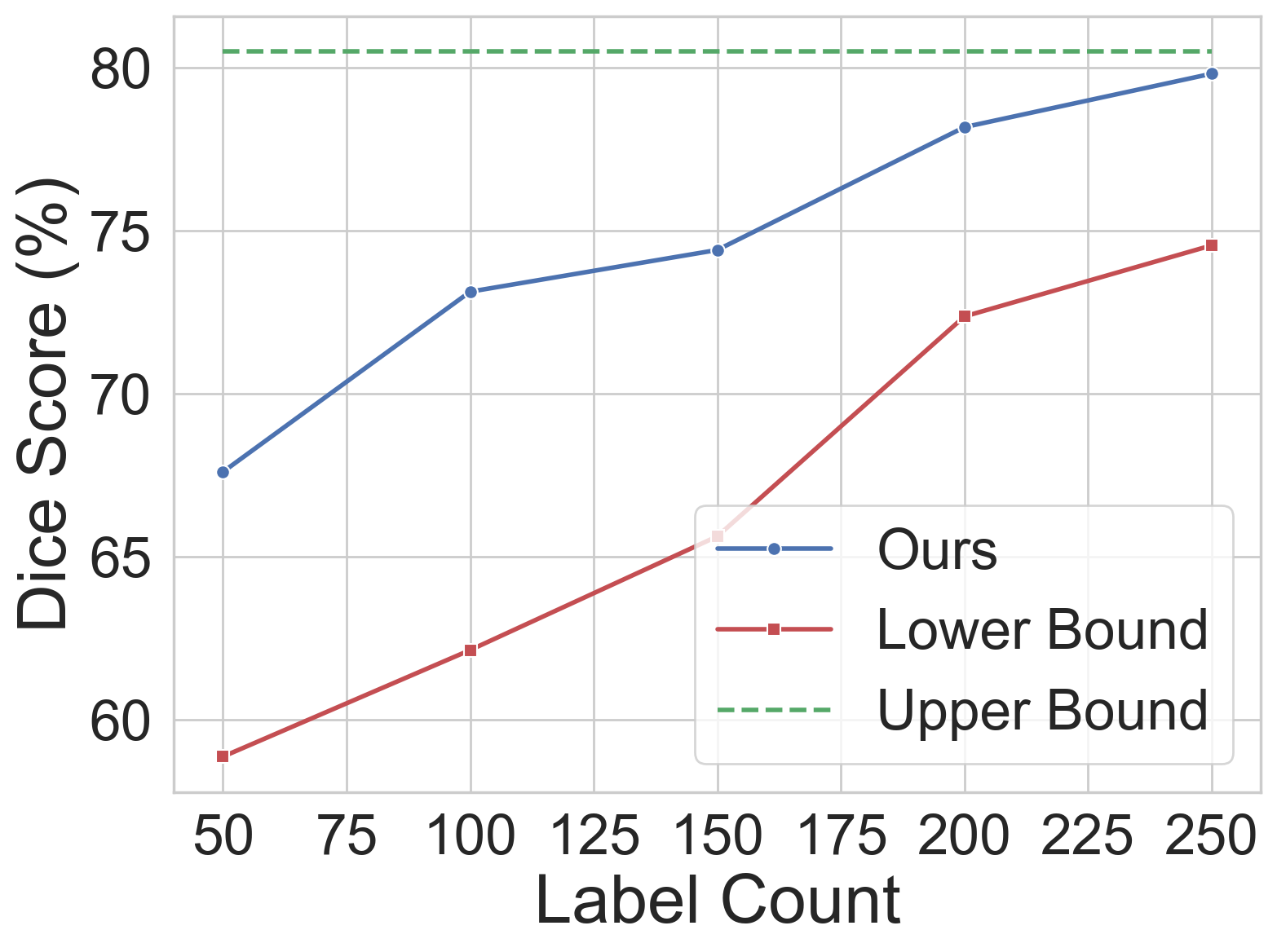} 
        \caption{Task 1}
    \end{subfigure}
    \hfill
    \begin{subfigure}[b]{0.32\textwidth}
        \centering
        \includegraphics[width=\textwidth]{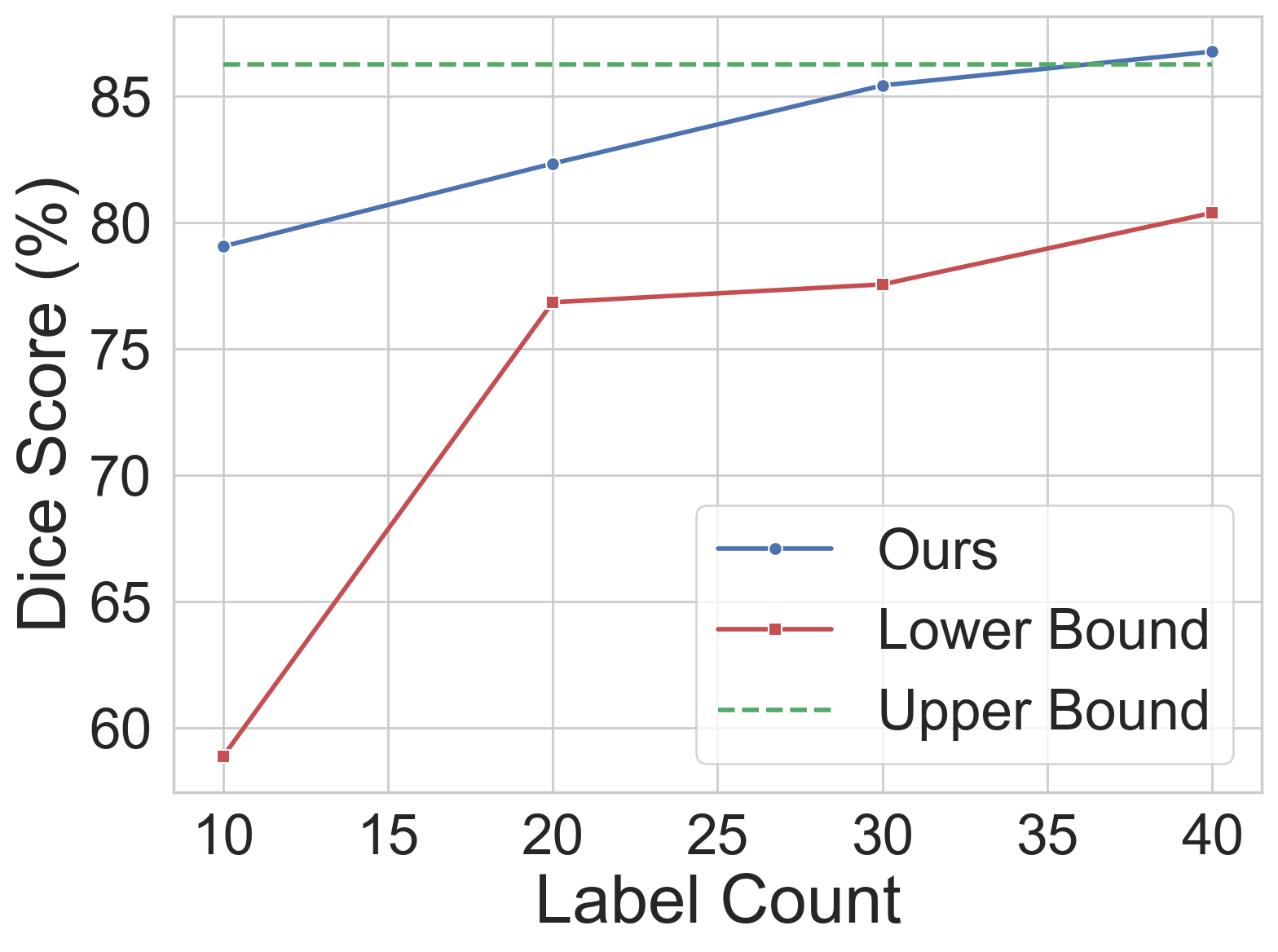} 
        \caption{Task 2}
    \end{subfigure}
    \hfill
    \begin{subfigure}[b]{0.32\textwidth}
        \centering
        \includegraphics[width=\textwidth]{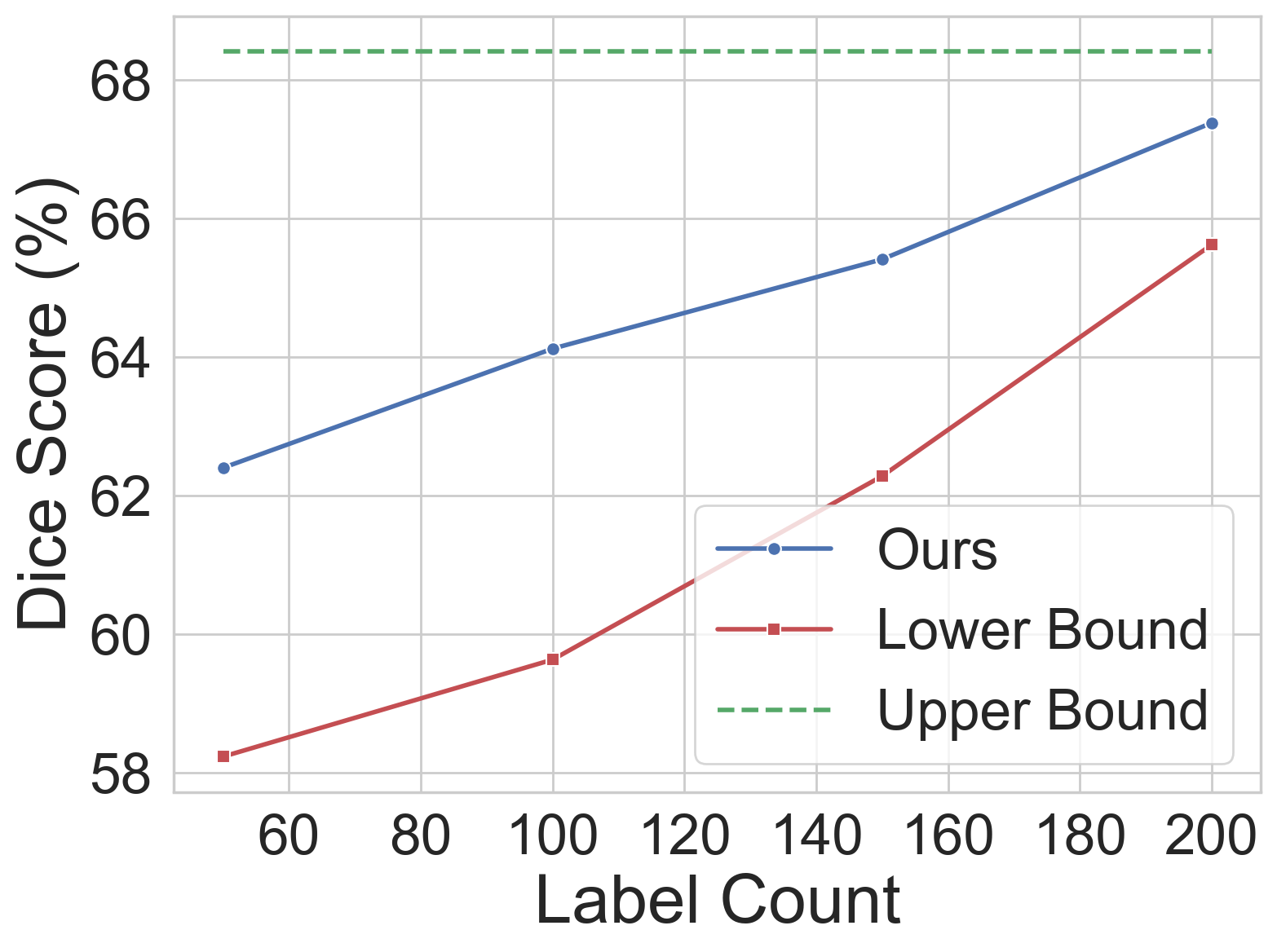} 
        \caption{Task 3}
    \end{subfigure}
    \caption{The relationship between the label count on each client and the average Dice score on Task 1: Cardiac MRI Segmentation, Task 2: Spine MRI Segmentation, and Task 3: Bladder Cancer Segmentation.}
    \label{fig5}
\end{figure}

\begin{table}[ht]
\centering
\caption{Ablation study on the effectiveness of each component in the proposed framework on Task 1: Cardiac MRI Segmentation. The best results are highlighted in bold.}
\begin{tabular}{ccc|cccc}
\toprule
GAA & DR  & PIA & Avg. DC (\%) $\uparrow$ & Avg. JC (\%) $\uparrow$ & Avg. HD95 (mm) $\downarrow$ & Avg. ASD (mm) $\downarrow$ \\ \midrule
& & & 60.05 & 51.92 & 32.49 & 27.11 \\
\checkmark & &  & 62.37 & 52.31 & 31.55 & 26.78 \\
& \checkmark &  & 61.89 & 51.67 & 30.72 & 25.93 \\
& & \checkmark  & 61.52 & 52.03 & 31.60 & 26.82 \\
\checkmark & \checkmark &  & 64.92 & 56.33 & 29.91 & 24.09 \\
\checkmark & & \checkmark  & 65.72 & 57.19 & 30.22 & 25.67 \\
& \checkmark & \checkmark  & 64.78 & 57.05 & 28.65 & 23.98 \\
\checkmark & \checkmark & \checkmark & \textbf{67.60} & \textbf{59.76} & \textbf{26.65} & \textbf{21.86} \\ \bottomrule
\end{tabular} \label{tab:ablation}
\end{table}

\subsection{Ablation Study}
\subsubsection{Effectiveness of Components}
We conducted an ablation study to evaluate the effectiveness of each component in our proposed framework as shown in Table \ref{tab:ablation}.  We applied FixMatch \citep{sohn2020fixmatch} as the SSL baseline and FedAvg \citep{mcmahan2017communication} as the FL model aggregation baseline. The introduction of the global aggregation-aware (GAA) component, dual-teacher refinement (DR), and perturbation-invariant alignment (PIA) individually improved the average Dice score (DC) by 2.37\%, 1.89\%, and 1.52\%, respectively, with GAA being slightly more effective than the other two components. The results indicate that each component contributes to the overall performance improvement. Additionally, we assessed the performance of various two-component combinations. All combinations achieved better performance than the individual components, with the GAA and DR combination achieving slightly better results. Finally, integrating all three components yielded the best performance, achieving an average Dice score of 67.60\%. These results highlight the importance of harmonizing global and local strategies within the FedSemiDG setting to achieve optimal outcomes.

\subsubsection{Effect of Label Count on Performance}
We performed a detailed analysis to examine how varying the label count in each domain, while ensuring it does not exceed the total number of labeled slices, impacts the performance of our method. The performance of our method ("Ours") is compared with the lower bound (FedAvg with only labeled data) and the upper bound (FedAvg with fully labeled data) on the three segmentation tasks (Task 1, Task 2 and Task 3). As shown in Fig. \ref{fig5}, as the label count increases, all tasks exhibit a clear trend of steady improvement in Dice score. The performance gap between our method and the lower bound remains significant, highlighting the effectiveness of our approach in mitigating label scarcity. And the close alignment with the upper bound further validates the capability of our method to achieve competitive performance even with limited labeled data. Additionally, as the label count reaches a certain level, the performance of our method is close to the upper bound as seen in Task 1 and Task 3 or even surpasses it as seen in Task 2. These results emphasize that the proposed method is highly effective in utilizing labeled data, narrowing the performance gap to the upper bound while maintaining substantial improvements over the baseline lower bound.

\subsubsection{Hyperparameter Sensitivity Analysis}

\begin{figure}[t]
    \centering
    \begin{subfigure}[b]{0.32\textwidth}
        \centering
        \includegraphics[height=4.7cm]{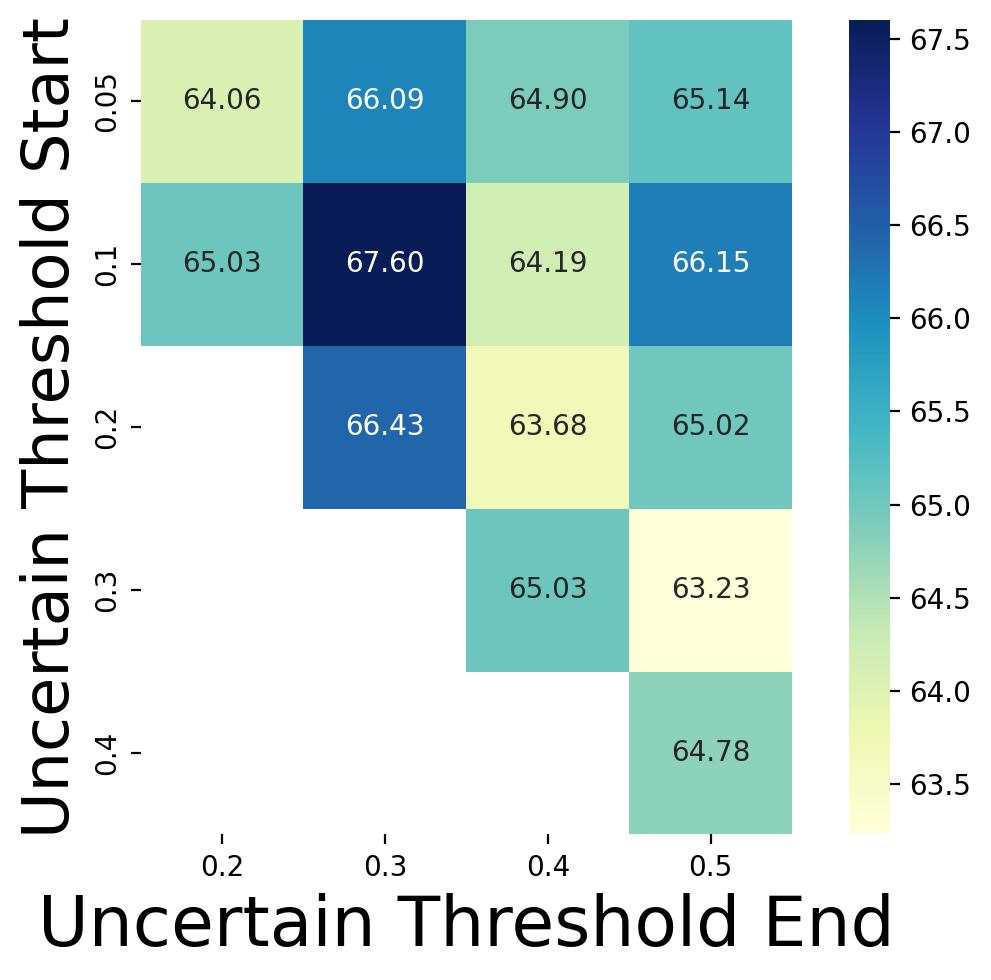} 
        \caption{Task 1}
    \end{subfigure}
    \hfill
    \begin{subfigure}[b]{0.32\textwidth}
        \centering
        \includegraphics[height=4.7cm]{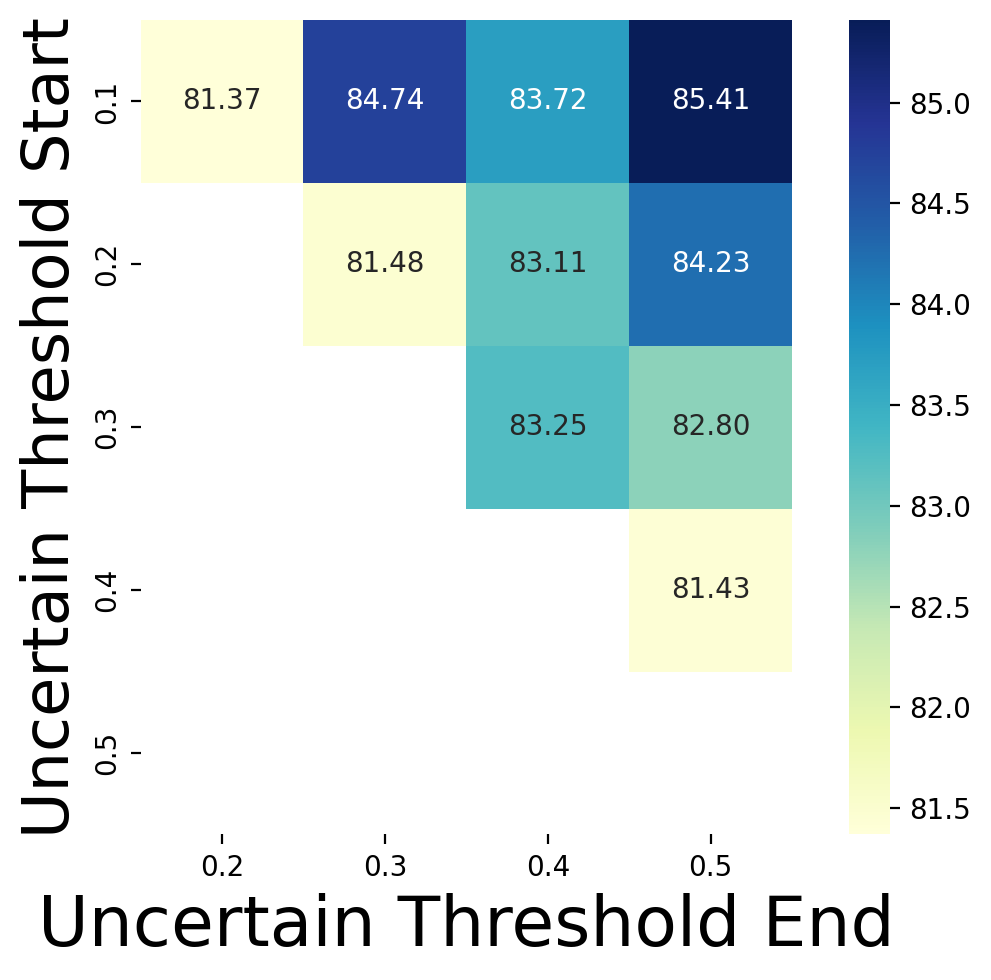} 
        \caption{Task 2}
    \end{subfigure}
    \hfill
    \begin{subfigure}[b]{0.32\textwidth}
        \centering
        \includegraphics[height=4.7cm]{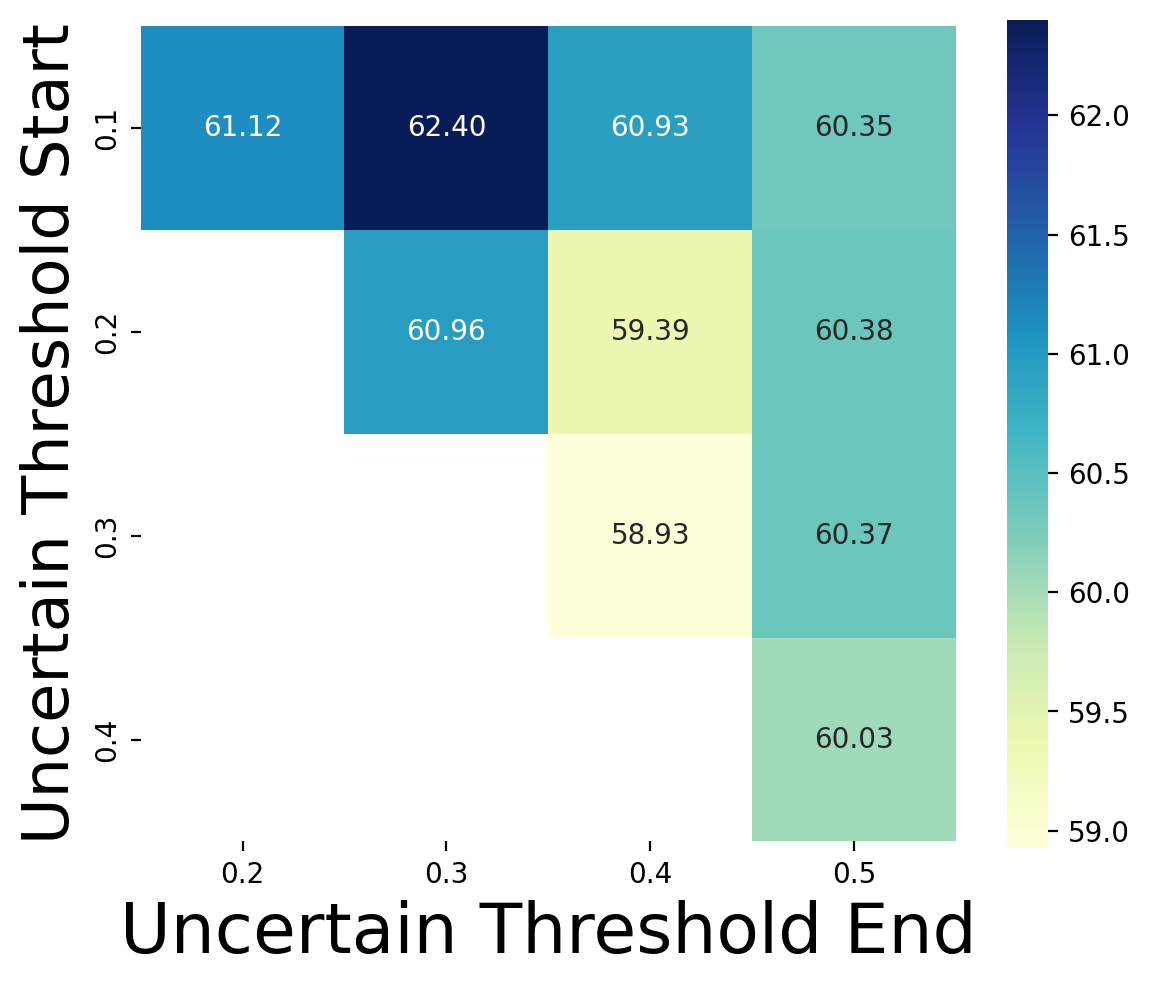} 
        \caption{Task 3}
    \end{subfigure}
    \caption{Hyperparameter analysis for the ramp-up quantile $\delta$ in the dual-teacher refinement on three segmentation tasks. $\delta$ is linearly increased from an initial value to a final value during the training process.The segmentation Dice scores (\%) are reported in the figure.}
    \label{fig6}
\end{figure}

\begin{figure}[t]
    \centering
    \begin{subfigure}[b]{0.32\textwidth}
        \centering
        \includegraphics[width=\textwidth]{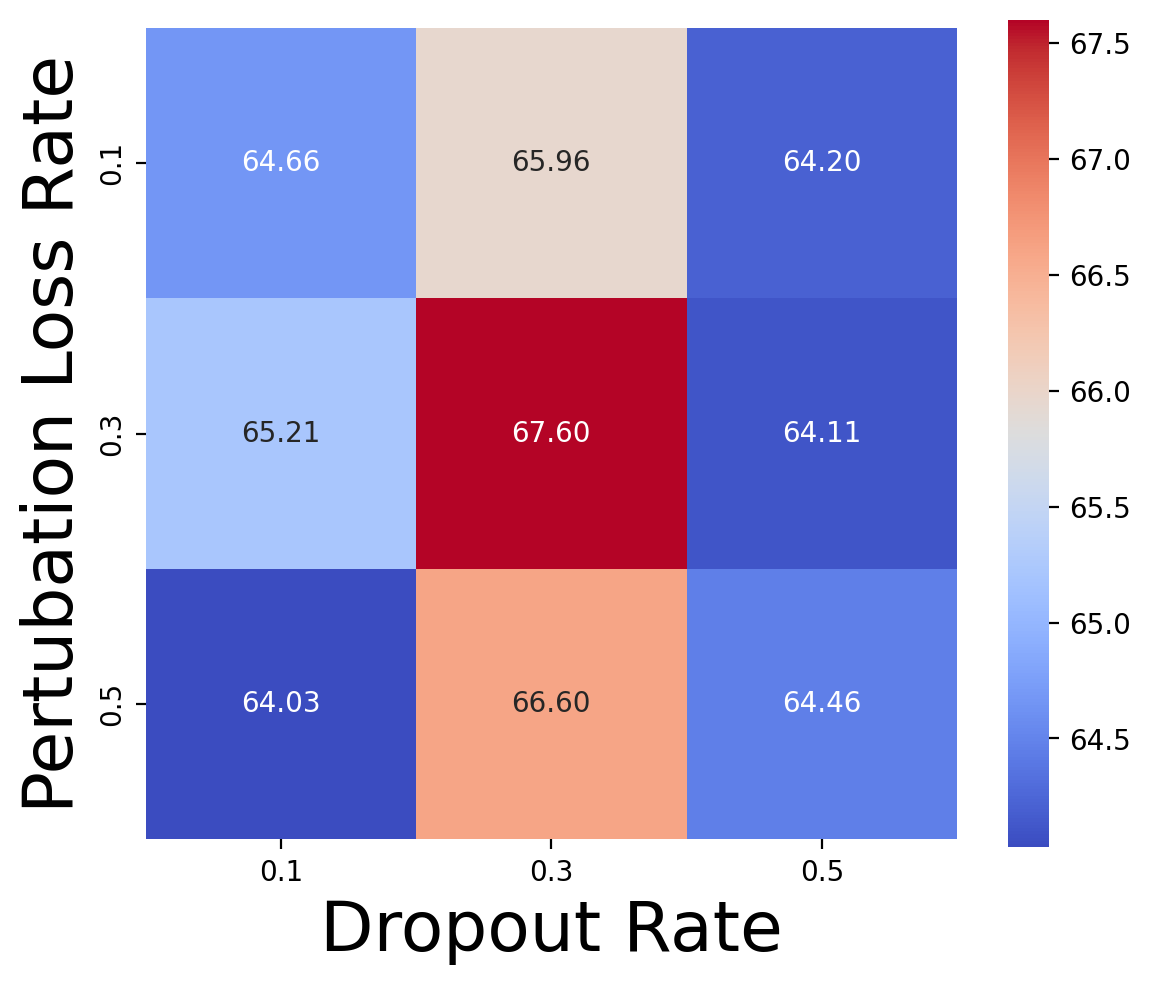} 
        \caption{Task 1}
    \end{subfigure}
    \hfill
    \begin{subfigure}[b]{0.32\textwidth}
        \centering
        \includegraphics[width=\textwidth]{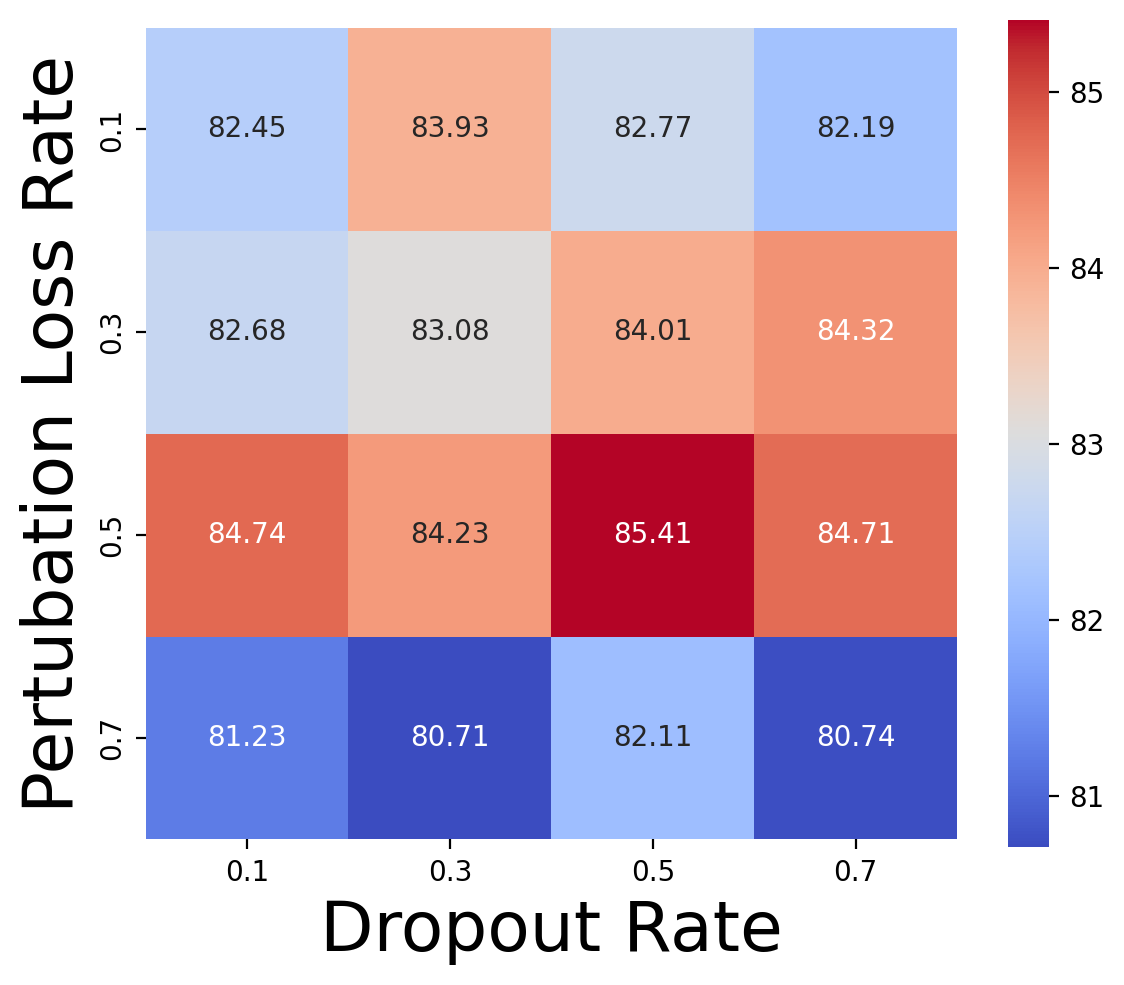} 
        \caption{Task 2}
    \end{subfigure}
    \hfill
    \begin{subfigure}[b]{0.32\textwidth}
        \centering
        \includegraphics[width=\textwidth]{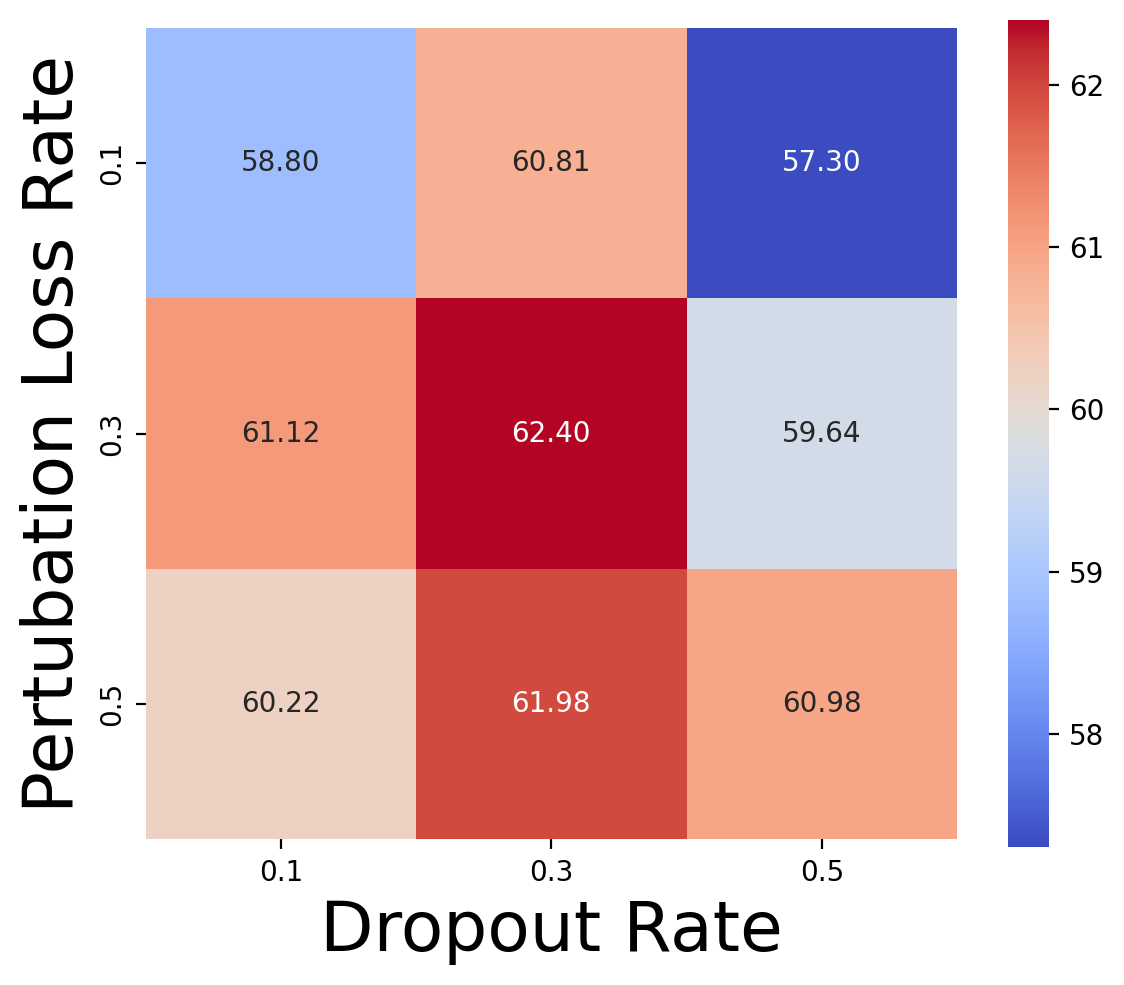} 
        \caption{Task 3}
    \end{subfigure}
    \caption{Hyperparameter analysis for the trade-off hyperparameter $\lambda_{2}$ in the perturbation-invariant alignment loss and dropout rate in the feature perturbation on three segmentation tasks.The segmentation Dice scores (\%) are reported in the figure.}
    \label{fig7}
\end{figure}

We conducted a detailed hyperparameter sensitivity analysis to evaluate the impact of two key components in our framework: the ramp-up quantile $\delta$ in the dual-teacher refinement process and the trade-off parameter $\lambda_2$ in the perturbation-invariant alignment loss combined with the feature dropout rate. These analyses, presented in Figs. \ref{fig6} and \ref{fig7}, provide insights into the sensitivity of our framework to these hyperparameters and identify configurations that optimize performance across three segmentation tasks.

For the ramp-up quantile $\delta$, we analyzed different combinations of starting and ending uncertainty thresholds. The results indicate that smaller starting values (0.1) paired with moderate ramp-up ranges (ending values of 0.3 to 0.4) consistently improve performance across tasks. This configuration balances the trade-off between exploiting confident predictions and refining uncertain ones during training, leading to stable and enhanced Dice scores.

In the analysis of the trade-off parameter $\lambda_2$ and the feature dropout rate, moderate values for both hyperparameters emerged as optimal. Specifically, setting $\lambda_2$ and the dropout rate to 0.3 resulted in the best performance for Task 1 and Task 3, while setting $\lambda_2$ and the dropout rate to 0.5 achieved the best performance for Task 2. Task 2 showed the most significant improvement under these settings, achieving a Dice core of 85.41\%. These findings highlight the effectiveness of our method in leveraging perturbation-invariant alignment to improve generalization.

Overall, these analyses emphasize the importance of proper hyperparameter tuning in maximizing the robustness and accuracy of our framework. The consistent trends across three tasks (Task 1, Task 2 and Task 3) suggest that the identified optimal ranges for $\delta$, $\lambda_2$, and dropout rate can serve as practical guidelines for similar segmentation tasks.

\subsubsection{Ablation Study on Feature Perturbation Strategies}
\label{sec:perturb}

\begin{figure}[t]
    \centering
    \begin{subfigure}[b]{0.32\textwidth}
        \centering
        \includegraphics[width=\textwidth]{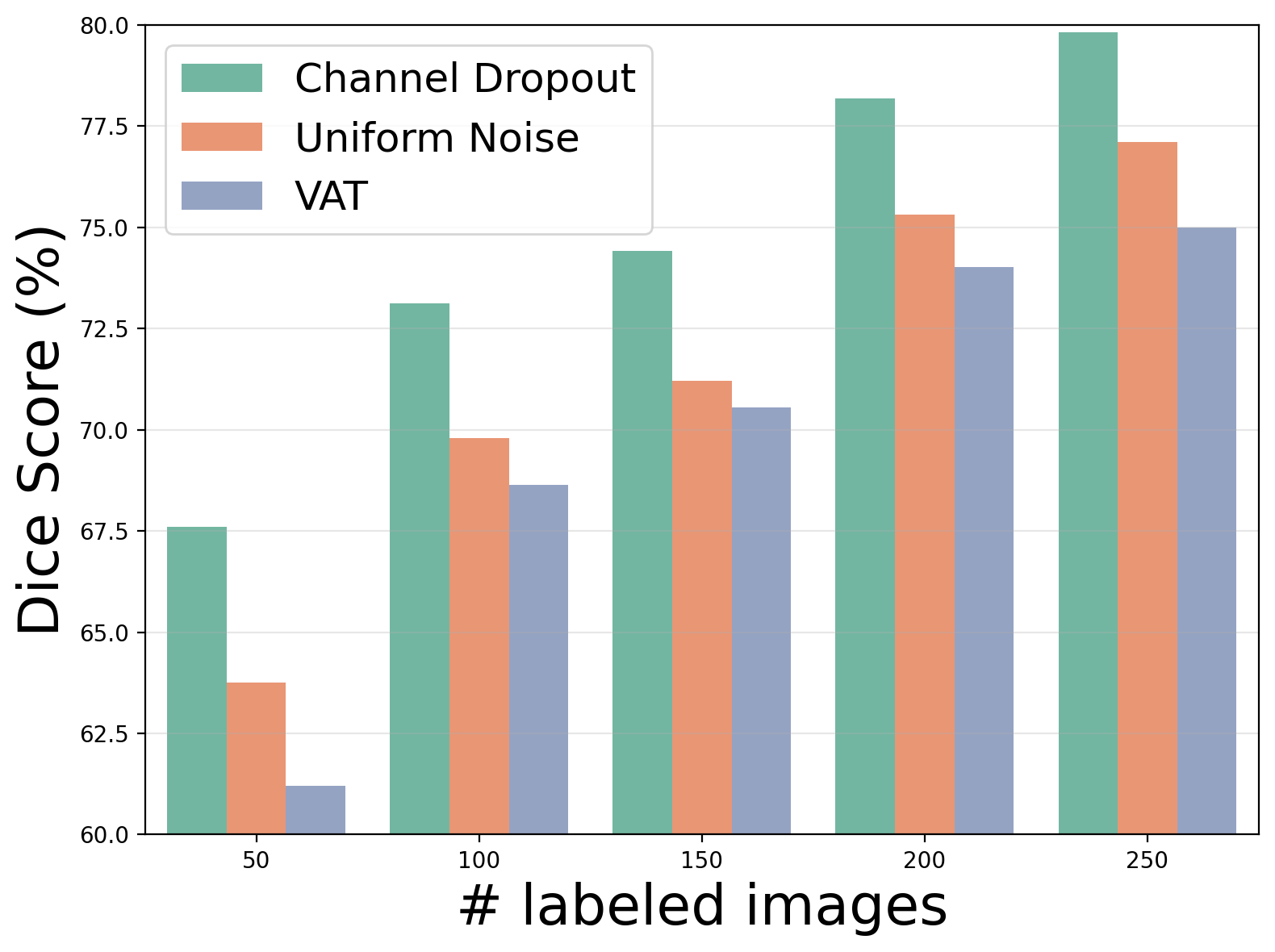} 
        \caption{Task 1}
    \end{subfigure}
    \hfill
    \begin{subfigure}[b]{0.32\textwidth}
        \centering
        \includegraphics[width=\textwidth]{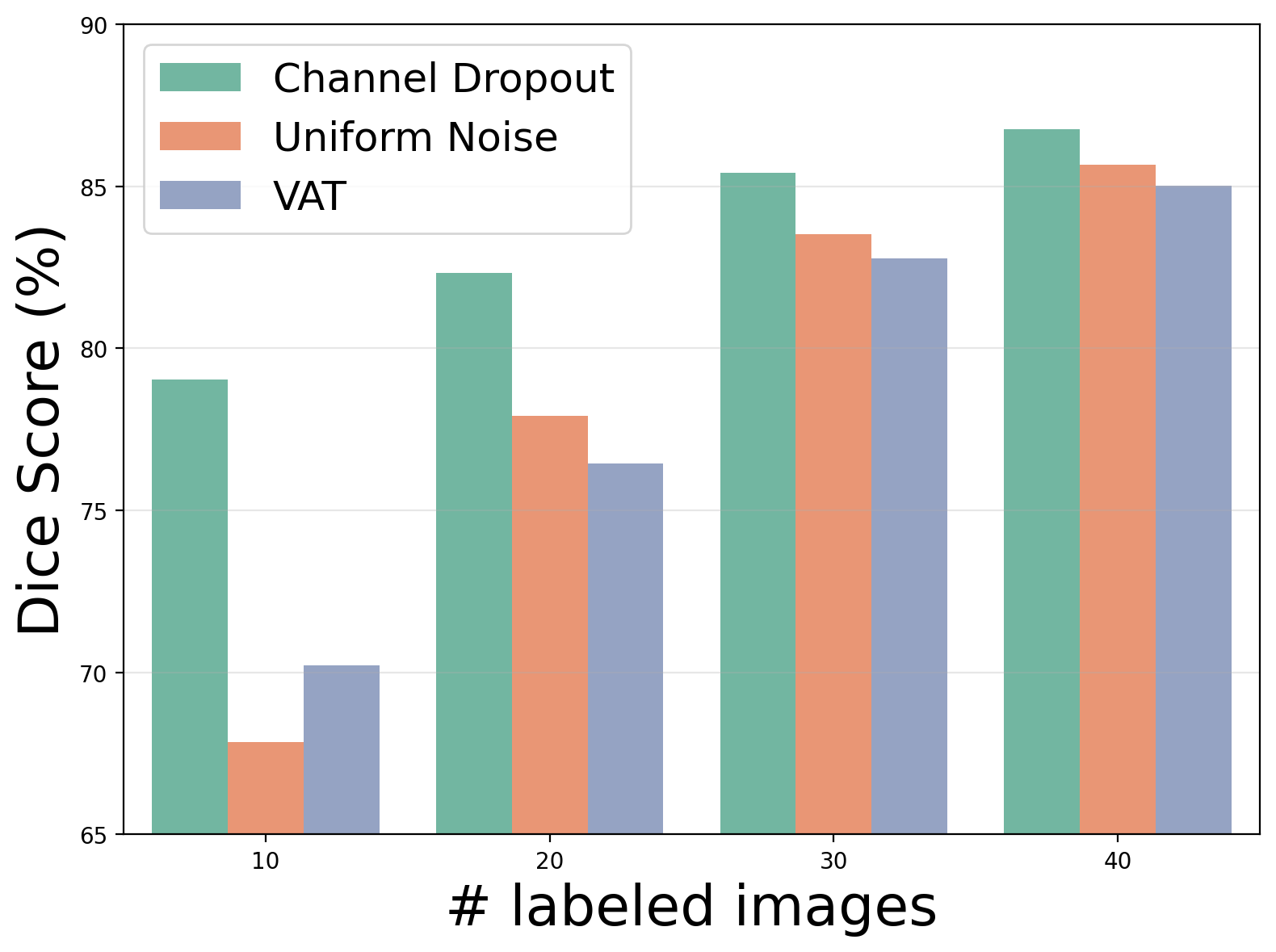} 
        \caption{Task 2}
    \end{subfigure}
    \hfill
    \begin{subfigure}[b]{0.32\textwidth}
        \centering
        \includegraphics[width=\textwidth]{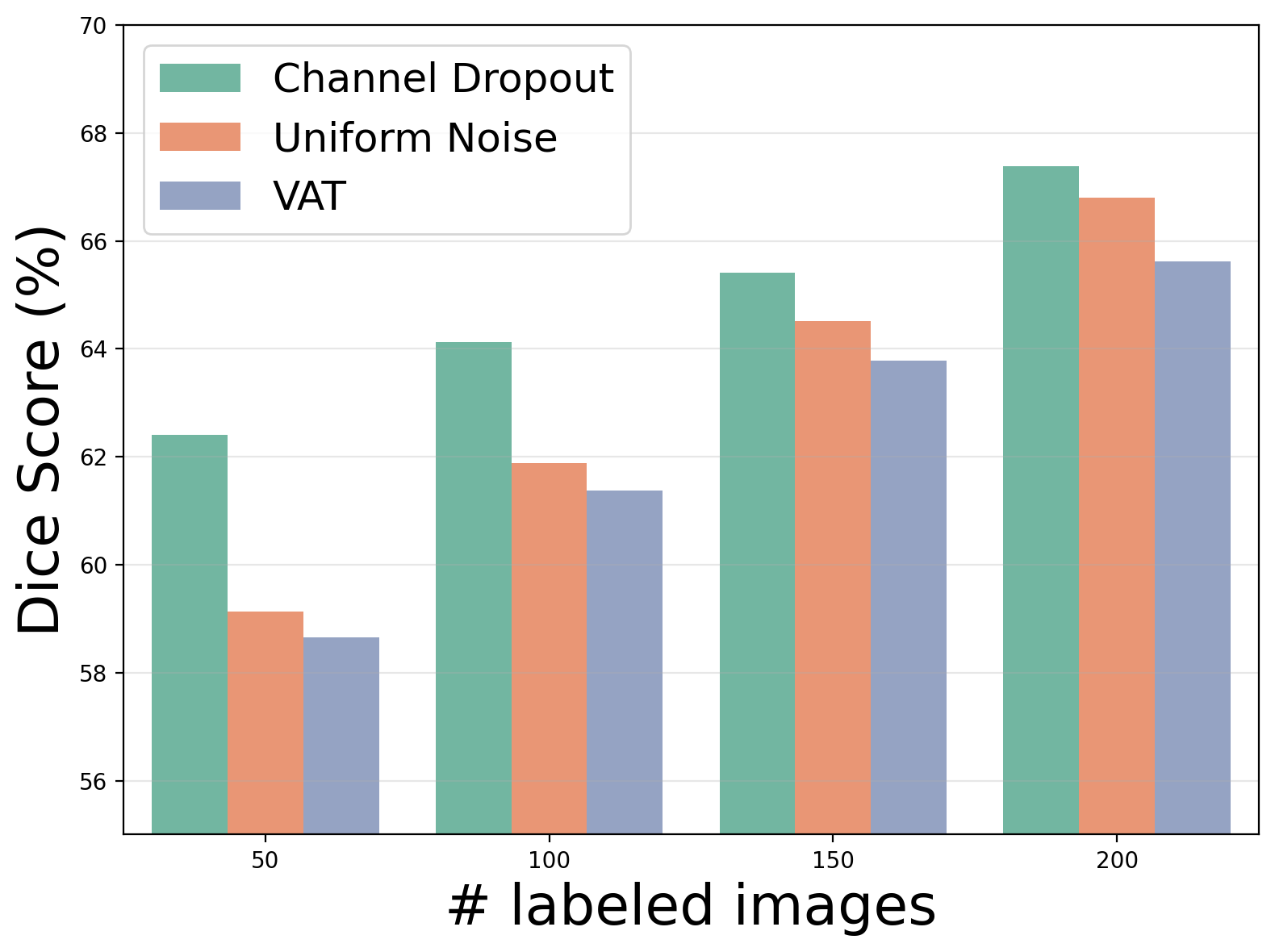} 
        \caption{Task 3}
    \end{subfigure}
    \caption{Ablation study on the efficacy of different feature perturbation strategies in our perturbation-invariant alignment component on three segmentation tasks.}
    \label{fig8}
\end{figure}

In our method, we utilize channel dropout as a straightforward and effective feature perturbation strategy. Alternative approaches, including uniform noise and virtual adversarial training (VAT) \citep{miyato2018virtual}, are also considered. To ensure a fair comparison, we adopt the hyperparameter settings outlined in \citep{ouali2020semi} for these strategies. Fig. \ref{fig8} presents an ablation study evaluating the efficacy of different feature perturbation strategies in the perturbation-invariant alignment (PIA) component across three segmentation tasks (Task 1, Task 2 and Task 3), with performance measured using Dice scores across varying numbers of training labeled images.

Across all three tasks, channel dropout consistently achieves the highest performance regardless of the number of labeled images. Its advantage is most pronounced at lower label counts, where it significantly outperforms uniform noise and VAT. As the number of labeled images increases, the performance gap between channel dropout and the other methods narrows slightly but remains consistent. Uniform noise shows competitive performance at higher label counts but falls short compared to channel dropout. VAT consistently performs the worst among the three strategies in all tasks. These results underline the effectiveness of channel dropout as a perturbation strategy within the PIA component. Its ability to introduce meaningful feature-level perturbations enhances generalization and segmentation accuracy across different tasks and label counts.

\subsubsection{Efficiency and Convergence Analysis}

We analyzed the efficiency of our framework from two perspectives: computational cost per round and convergence speed over multiple rounds.

\begin{figure}[t]
    \centering
    \begin{subfigure}[b]{0.48\textwidth}
        \centering
        \includegraphics[width=\textwidth]{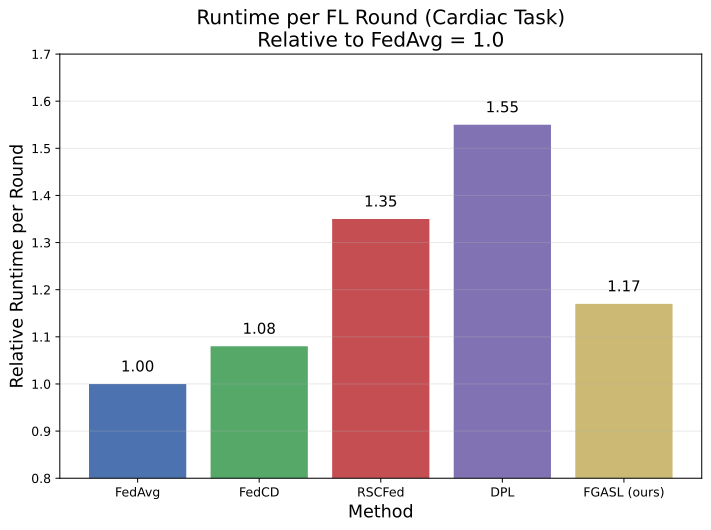} 
        \caption{Runtime per FL round, normalized by FedAvg.}
        \label{fig:runtime_bar}
    \end{subfigure}
    \hfill
    \begin{subfigure}[b]{0.48\textwidth}
        \centering
        \includegraphics[width=\textwidth]{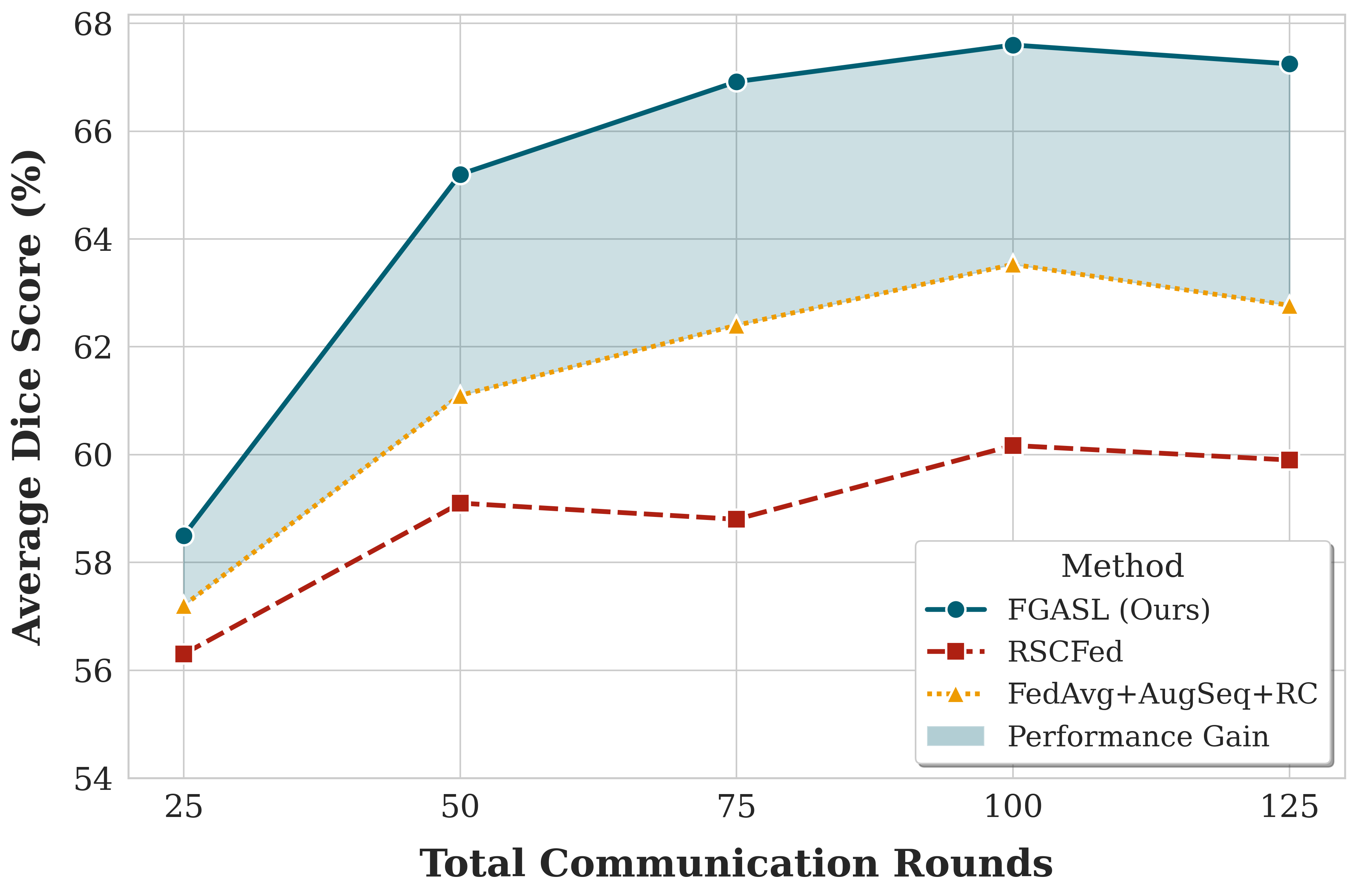} 
        \caption{Convergence speed vs. total communication rounds.}
        \label{fig:convergence_line}
    \end{subfigure}
    \caption{Efficiency and performance analysis on the Cardiac MRI Task. (a) Relative computational cost per round. (b) Segmentation performance achieved at various total communication rounds.}
    \label{fig:efficiency_analysis}
\end{figure}

\noindent \textbf{Computational Efficiency.} We first analyzed the runtime efficiency by measuring the time per federated round on the Cardiac MRI segmentation task. Fig.~\ref{fig:efficiency_analysis}(a) presents the relative per-round runtime, normalized by FedAvg as the baseline. Our method FGASL introduces a modest runtime increase of about 17\% compared to FedAvg, which is primarily due to the dual-teacher forward passes. In contrast, other methods like DPL incur higher overhead due to repeated Monte Carlo dropout passes. This demonstrates that FGASL achieves a favorable balance between performance and computational cost per round.

\noindent \textbf{Convergence and Communication Efficiency.}
To analyze the sensitivity to the number of communication rounds, we conducted an additional experiment on the Cardiac MRI segmentation task. We trained both our model and the baseline from scratch for varying total communication rounds. As shown in Fig.~\ref{fig:efficiency_analysis}(b), our FGASL consistently outperforms the baseline across all communication budgets and the performance advantage of FGASL over baselines widens as training progresses. We provide an intuitive analysis for this observation. In the early training phase, the models are not yet powerful. Consequently, the estimated Generalization Gap is a noisy and less reliable signal for guiding aggregation, limiting the immediate impact of our Generalization-Aware Aggregation (GAA) module. However, as the models mature with continued training, the gap becomes a more stable and meaningful indicator of how the global model fits each domain. This allows our GAA to more effectively up-weight under-represented domains, leading to a compounding improvement in generalization as the training continues. This superior learning dynamic is evident in that FGASL surpasses the best baseline's peak Dice score (at 100 rounds) in just 50 rounds.

\section{Discussion}

Collecting large amounts of labeled medical data from a single center is inherently challenging, making Federated Semi-Supervised Learning (FSSL) an attractive solution for medical image segmentation tasks. Multi-center collaboration under the FSSL paradigm is critical for advancing medical image analysis, as it leverages diverse data from multiple sources and enables the development of robust models that generalize well, all while preserving data privacy. However, existing FSSL methods often overlook the pervasive issue of domain shifts in multi-center medical imaging data. Our work addressed this gap by introducing a novel framework tailored for Domain Generalized Federated Semi-Supervised Learning (FedSemiDG), a setting that has been underexplored yet is highly relevant to real-world medical imaging scenarios. 

Technically, our experiments showed that previous FSSL methods only brought marginal improvements over the FL lower bound, while our approach demonstrated strong performance in this challenging scenario, demonstrating the effectiveness of our proposed framework. Clinically, FedSemiDG is particularly relevant for real-world multi-center medical imaging applications, where data is often scarce and domain shifts are prevalent. Our experiments shows that local training could not achieve satisfactory performance due to the lack of access to multi-center data even with SSL methods, while our approach significantly improved the performance, showing great potential for clinical applications.

\noindent \textbf{Limitations and Future Work.}
Despite the promising results, our work has several limitations that suggest directions for future research. Firstly, our FGASL framework introduces a modest computational overhead compared to vanilla FedAvg. Future work could explore more lightweight aggregation and refinement strategies to improve efficiency. Secondly, while our framework is inherently privacy-preserving by only transmitting model weights and a single scalar gap value, more rigorous privacy protections could be established by incorporating advanced cryptographic techniques. For instance, Differential Privacy \citep{abadi2016deep} could be applied to the shared model parameters, and Secure Aggregation \citep{bonawitz2017practical} could be used to protect the individual gap values during the computation of their mean, making our framework robust against even stronger adversarial attacks.

Furthermore, integrating foundation models such as MedSAM \citep{ma2024segment} presents a promising direction. As suggested by \citep{ma2025steady}, while these models may produce overconfident predictions, they can serve as powerful pseudo-labelers or initialization backbones. Investigating how to best leverage them within a federated pipeline to balance generalisation, adaptability, and communication efficiency is a promising research avenue. Moreover, extending our framework to more complex scenarios, such as those with severely imbalanced datasets or where some clients are entirely unlabeled, remains an important direction for future exploration. Future research could also include benchmarking the proposed framework on other imaging modalities (e.g., CT, ultrasound) and tasks (e.g., classification, detection), to further assess its generalization capability.

\section{Conclusion}
In this paper, we tackled the under-explored problem of FedSemiDG. Our proposed framework FGASL, which integrates global and local strategies, achieved robust generalization across unseen domains in challenging medical image segmentation tasks. These results highlighted the promise of FedSemiDG for advancing federated learning applications in healthcare.




\printcredits

\bibliographystyle{cas-model2-names}

\bibliography{cas-refs}



\end{document}